# Clue-Guided Money Laundering Group Discovery

Boyang Wang, Jianing Cao

*Abstract*—**Money Laundering Group Discovery (MLGD) aims to identify hidden criminal groups and recover their complete structures in large-scale financial networks. Existing graph anomaly detection methods mainly produce node-level risk alerts, while global group discovery methods passively search for suspicious groups over the whole network. Both are mismatched with real Anti-money-laundering (AML) investigations, where analysts usually start from a concrete clue and gradually expand the investigation to recover the responsible group. To address this gap, we propose Clue-Guided Group Discovery (CGGD), where a laundering group is progressively recovered from an initial clue set through analyst interaction. We further propose Clue2Group, a framework that first constructs a compact local investigation context to reduce noise and preserve chain-like and cycle-like laundering structures. It then estimates a clue-conditioned local risk field with a multi-semantic local-temporal GNN, and finally integrates risk, structural, and prior-pattern evidence to recover a coherent laundering group. Experiments on two large-scale AML benchmarks show that Clue2Group provides a practical clue-driven analysis framework for AML investigations, offering a feasible step toward bridging the gap between graph-based AML research and real investigation workflows.**

## I. INTRODUCTION

MONEY laundering is a major financial crime that continues to threaten the stability of the global financial system. With the rapid development of financial technologies and cross-border payment networks, laundering operations have shifted from isolated accounts to complex transaction networks that enable multi-stage fund layering and risk dispersion [1-3]. These graph-structured criminal activities are highly concealed, pushing anti-money-laundering (AML) research toward identifying hidden criminal organizations in financial networks, known as Money Laundering Group Discovery (MLGD). In practice, investigations follow a clue-driven, analyst-in-the-loop process [4, 5]: starting from one or a few initial clues, analysts iteratively expand the investigation to recover the full laundering group most relevant to the given clues. The key objective is therefore conditional and targeted structural recovery centered on observed clues.

Under this workflow, it is important to clarify the logical relationship between Graph-based Financial Anomaly Detection (GFAD) and MLGD in the anti-money-laundering pipeline. GFAD methods [6-8] serve the alerting stage: they assign node- or transaction-level risk scores across the full network to generate candidate clues, but do not answer *what group structure lies behind a given clue?*

In contrast, MLGD should be positioned as a post-alert, investigation-stage task [9]. Global-view MLGD methods [10-12] go further by discovering candidate groups through subgraph mining or community detection, yet their optimization targets network-wide coverage rather than clue-specific recovery, leaving analysts to manually associate outputs with a particular case. Recent work has begun to explore local sub-network construction around suspicious entities [13], but these local views are typically built once as auxiliary detection units rather than modeled as clue-conditioned contexts that evolve with the investigation. In short, no existing studies directly supports conditional, clue-centered group recovery at the investigation stage.

Inspired by interactive community search[14, 15], we formulate **Clue-Guided Group Discovery (CGGD)**, an investigation-stage task that casts MLGD as a conditional group recovery problem. Given a background transaction graph and initial clues, the model outputs a group subgraph tightly associated with that clue. Unlike one-shot global discovery, CGGD treats group recovery as an iterative, analyst-coupled process that directly supports investigative decision-making.

CGGD faces two main challenges, and our proposed **Clue2Group** framework is designed to address them. First, initial clues are often extremely sparse in a large noisy graph, making it difficult to directly locate an effective investigation region. To address this issue, **Structure-Aware Context Construction (SACC)** constructs a compact and structurally enriched local investigation context. This helps narrow the search space while preserving key structures related to potential laundering groups. Second, clues usually cover only a small number of group members. Clue2Group therefore needs to identify which nodes may have collaborative relationships with the given clues and further recover the corresponding group boundary. To address this challenge, **Multi-Semantic Local-Temporal Graph Neural Network (MIST-GNN)** estimates a conditional risk field by modeling local temporal context and multi-semantic relational signals. **Evidence-Driven Group Assembly (EDGA)** then integrates risk, structural, and prior-pattern evidence to recover a clue-relevant and structurally coherent laundering group.

In summary, the main contributions of this article are as follows:

- We propose the task of CGGD. This task shifts money laundering group discovery from traditional global anomaly detection or global group mining to clue-conditioned group recovery at the investigation stage.

Boyang Wang is with the Department of Informatics, King's College London, WC2B 4BG London, U.K. (e-mail: boyang.2.wang@kcl.ac.uk / wwby2502@163.com ).

Jianing Cao is with the School of Transportation, Southeast University, Nanjing, 211189, China. (e-mail: jianing.cao@connect.polyu.hk )

- We propose and formalize the Clue2Group framework. The framework decomposes CGGD into three stages: Clue-Guided Context Construction, Conditional Risk-Field Estimation and Group Assembly. This design supports the progressive recovery of a relevant group boundary from sparse clues.
- We provide a systematic empirical study of Clue2Group on two large-scale AML benchmarks. The results verify its effectiveness in context construction, risk estimation, group assembly, and end-to-end interactive recovery. Ablation, sensitivity, efficiency, and case studies further show that the framework is robust, computationally practical, and able to provide interpretable evidence for AML investigations.

## II. RELATED WORK

### A. Graph-based Financial Fraud Detection

AML techniques have evolved from rule-based systems [16, 17] to machine-learning classifiers over handcrafted features [6, 18] and, more recently, to graph mining and graph neural networks that operate on global transaction networks [7, 8, 19-21]. Each generation has improved alerting accuracy, yet these methods fundamentally produce node- or transaction-level risk scores for alert generation, rather than group-level structures that can be directly used for case-level investigation.

### B. Global-View Money Laundering Group Discovery

Beyond node-level anomaly detection, some studies identify money laundering groups from a global perspective through subgraph mining, collective anomaly detection, or community detection [10-12]. These approaches are effective for broad-coverage macro-level risk analysis, but their optimization targets network-wide partition quality, producing candidate group sets that are not conditioned on specific investigative clues.

### C. Target-Conditioned Local Subgraph Discovery

A closely related line of work is learning-based community search[22]. Building on this line, ICS-GNN[14] introduces an interactive mechanism. It allows users to refine the results by labeling positive and negative examples. In this sense, it establishes a basic form of interactive and query-conditioned target subgraph discovery.

However, this paradigm is mainly designed for preference-driven community search and differs from AML group recovery in two key aspects. First, it treats a community as a similarity cluster defined by user preferences, rather than as an objective entity. Thus, the same query vertex may produce different communities. In AML investigations, the target laundering group is an objective criminal collaboration entity. The analyst's clues are only partial observations of this entity, and the goal is to approximate its true boundary, not to construct a preference-matched community.

Second, this paradigm often equates model scores with the community definition itself. For example, ICS-GNN directly defines the target community as the connected subgraph with the maximum sum of GNN scores. This is not sufficient when the goal is to recover an objective entity. No single source of evidence can reliably determine its boundary[23, 24]. Therefore, in Clue2Group, model scores are not treated as the decisive factor for membership. Instead, they serve as one source of evidence, together with structural evidence and prior-pattern evidence. The final group boundary is obtained through cross-validation among multiple independent evidence sources.

Relatedly, recent studies have constructed local transaction sub-networks around suspicious entities for risk assessment [13], and GNN explainers [25-27] generate explanatory subgraphs conditioned on individual predictions. However, explainers optimize for faithful explanation of model outputs rather than for reconstructing laundering processes and recovering group boundaries, limiting their applicability to case-level group tracing.

### D. Heterogeneous and Temporal Graph Learning in Financial Networks

Financial transaction networks are inherently heterogeneous and temporal, and a rich body of methods exploits these properties for fraud and anomaly detection [28, 29]. However, most approaches train over the global network, incurring substantial computational overhead that conflicts with the interactive computation required at the investigation stage. Under our clue-driven setting, we instead construct a compact local context and introduce heterogeneous and temporal semantics within it, leveraging modeling power at controlled cost.

## III. PROBLEM DEFINITION

### A. Preliminaries

*Definition 1. Financial Transaction Network*

We model a financial network as a dynamic directed multigraph $G=(V,E)$, where $V$ denotes the set of account nodes, and $E=\{e_i\}_{i=1}^{m}$ is a multiset of transaction events, allowing multiple transactions between the same pair of accounts. Each transaction event is represented as:

$$e_i=\left(u_i,v_i,t_i,x_i\right) \quad u_i,v_i \in V, t_i \in T, x_i \in \mathbb{R}^{d_e}, \tag{1}$$

where $u_i$ $u_i$ and $v_i$ denote the source and destination accounts, $t_i$ is the timestamp drawn from an ordered time domain $T$, and $x_i$ is a transaction attribute vector (e.g., amount, currency, payment type).

*Definition 2. Investigative Clues*

We define the investigative clues as a set of transaction events that have been flagged as suspicious $C_E \subseteq E$. The clue set induces a node-level clue set:

$$C_V=\left\{u,v \middle| e=\left(u,v,t,\mathrm{x}_e\right) \in C_E\right\}. \tag{2}$$

Let its connected components partition $C_V$ as:

$$C_V=C_V^{(1)} \cup C_V^{(2)} \cup \cdots \cup C_V^{(r)}, \quad r \geq 1. \tag{3}$$

Each $C_V^{(r)}$ is called a clue component. We assume the clues are triggered by one dominant laundering group. At the initial stage $r=1$ and $C_V$ is connected; during interaction, $r$ may grow as analyst-confirmed positives are added to $C_V$, with each component serving as an additional anchor on the same target group.

*Definition 3. Induced Group Subgraph*

Since edge selection is ambiguous in multigraphs, we use the node set as the primary decision variable for group hypotheses. Given any node set $Y \subseteq V$, the transaction-induced subgraph on $G$ is defined as:

$$E[Y] = \{e = \{u, v, t, x\} \in E \mid u, v \in Y\},$$
$$G[Y] = (Y, E[Y]). \tag{4}$$

Accordingly, all group subgraphs returned in this work are of the form $G[Y]$, where $Y$ denotes the inferred set of group member accounts.

*B. Clue-Guided Group Discovery*

Given a financial graph $G$, a node-level clue set $C_V$ and an analyst-verified benign node set $B_V$, we assume the true laundering group resides in a clue-related local context rather than the entire network. Accordingly, CGGD is modeled as a staged process with three components:

**Clue-Guided Context Construction**

$$\mathcal{G}_{LIC} = \Psi(G, C_V), \quad \mathcal{G}_{LIC} \subseteq G, \tag{5}$$

where $\Psi(\bullet)$ is a clue-guided context constructor that extracts a local subgraph $\mathcal{G}_{LIC} = (\mathcal{V}_{LIC}, \mathcal{E}_{LIC})$, with $C_V \subseteq \mathcal{V}_{LIC} \subseteq V$.

**Conditional Risk-Field Estimation**

$$p(v) = f_\Theta(v \mid \mathcal{G}_{LIC}, C_V, B_V) \in [0,1]^{|\mathcal{V}_{LIC}|}, \quad \mathcal{G}_{S-LIC} = (\mathcal{V}_{LIC}, \mathcal{E}_{LIC}, \mathbf{p}), \tag{6}$$

where $p(v)$ indicates the confidence that node $v$ belongs to the same laundering group as the clues. In the initial stage, $B_V = \varnothing$ and (6) is a PU learning task; in the interactive stage, the analyst provides $B_V \neq \varnothing$, promoting (6) to a PNU semi-supervised task.

**Group Assembly**

Each clue component $C_V^{(r)}$ is required to be covered by its own group hypothesis $Y^{(r)} \subseteq \mathcal{V}_{S-LIC}$ ,i.e., $C_V^{(r)} \subseteq Y^{(r)}$. The inference objective is:

$$Y^{(r)*} = \arg\max_{Y \subseteq \mathcal{V}_{S-LIC}} \Omega\left(Y \mid \mathcal{G}_{S-LIC}, C_V^{(r)}, \mathbf{p}\right), \quad \mathbf{p} = \{p(v)\}_{v \in \mathcal{V}_{S-LIC}}, \tag{7}$$

and the output group is the induced subgraph $\mathcal{G}_{S-LIC}\left[Y^{(r)*}\right]$.

## IV. Clue2Group

Fig. 1 gives an overview of the proposed Clue2Group framework. Starting from sparse investigative clues, Clue2Group first constructs a Local Investigation Context (LIC) using SACC, then estimates clue-conditioned risk scores with MIST-GNN, and finally assembles the target laundering group using EDGA by integrating risk, structural, and prior-pattern evidence. As detailed in the following subsections.

*A. Structure-Aware Context Construction (SACC)*

Given a financial network $G = (V, E)$ and a clue account set $C_V \subseteq V$, SACC constructs a Local Investigation Context (LIC) subgraph that satisfies three requirements:

- **Relevance**. The LIC should focus on potential group structures around the clues, suppressing background noise and irrelevant transactions.
- **Structure coverage**. The LIC should not only include nearby neighborhoods, but also proactively recover structural evidence commonly observed in laundering processes (e.g., chain- and cycle-like patterns) [10], reducing the risk of structural fragmentation in later assembly.
- **Controllability**. In a directed transaction multigraph, a single account pair may have a large number of transaction events. LIC construction therefore needs explicit controls on key parameters and computational cost to support analyst-in-the-loop iterations.

*Neighborhood Expansion*

For each clue node $c \in C_V$, we perform a k-hop neighborhood search on $G$ and obtain the candidate node set:

$$\mathcal{V}_{k-hop}(c) = \{v \in V \mid dist_G(c, v) \leq k\}. \tag{8}$$

*Structural Enhancement*

We define a topological path as $P_c = \langle v_0, \ldots, v_L \rangle$ with $v_0 = c$, where $L$ is the path length. For chain, we only enumerate simple paths. For cycle, we require the path to be closed:

$$P_c = \begin{cases} P_{cycle} = \langle v_0, \ldots, v_L \rangle, v_0 = c, v_L = v_0, \\ P_{chain} = \langle v_0, \ldots, v_L \rangle, v_0 = c, v_L \neq v_0. \end{cases} \tag{9}$$

To capture typical laundering chains while avoiding combinatorial explosion from overly long paths, we constrain chain lengths by $L \in [L_{\min}, L_{\max}]$, and impose a maximum length constraint $C_{\max}$ for cycles.

In investigations, the key question is whether there exists a self-consistent fund-flow explanation along a path, rather than requiring all parallel transactions to satisfy strict consistency. We therefore define the set of fund-flow instances along $P_c$ as:

$$\mathcal{I}(P_c) = \left\{ i \mid \forall \ell \in \{1, \ldots, L\}, e_\ell^{(i)} \in \mathcal{E}(v_{\ell-1}, v_\ell) \right\}, \tag{10}$$

where $\mathcal{E}(u, v) \subseteq E$ denotes the set of all transaction events between account pair $(u, v)$. Each instance $i \in \mathcal{I}(P_c)$ selects one concrete transaction event per hop, forming an event sequence $\langle e_1^{(i)}, \ldots, e_L^{(i)} \rangle$ .To keep the procedure controllable in multigraphs, we cap the number of candidate events per account pair: for any $(u, v)$, we retain at most $m_{\max}$ events $\tilde{\mathcal{E}}(u, v) \subseteq \mathcal{E}(u, v)$ and use them to construct $\mathcal{I}(P_c)$. This yields the bound $|\mathcal{I}(P_c)| \leq (m_{\max})^L$.

For each fund-flow instance $i \in \mathcal{I}(P_c)$, we extract its amount sequence $a^{(i)} = \langle a(e_1^{(i)}), \ldots, a(e_L^{(i)}) \rangle$, where $a(e)$ is the transaction amount of event $e$. We measure amount consistency using the coefficient of variation:

$$CV(a^{(i)}) = \frac{\sigma(a^{(i)})}{\mu(a^{(i)})}, \tag{11}$$

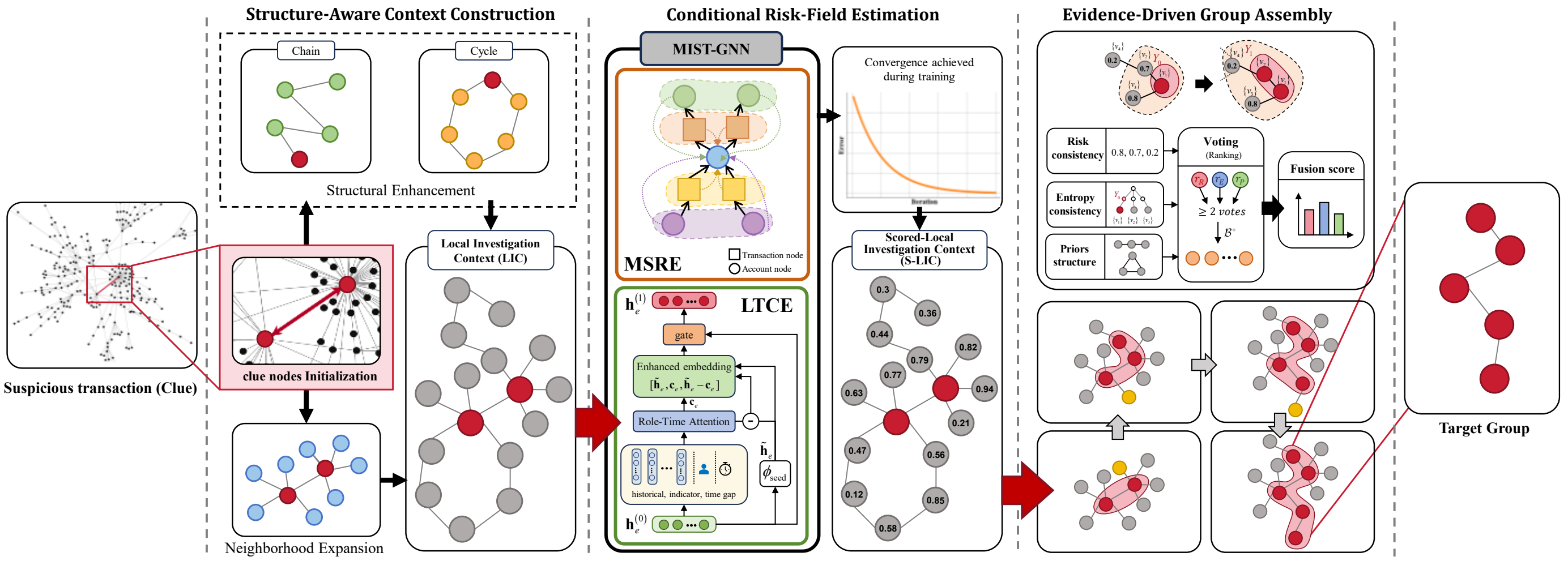

**Fig. 1.** Overview of the Clue2Group framework, which recovers a clue-relevant laundering group through three stages: context construction (SACC), conditional risk-field estimation (MIST-GNN), and evidence-driven group assembly (EDGA).

where $\sigma(\bullet)$ and $\mu(\bullet)$ denote the standard deviation and mean, respectively. A path $P_c$ is considered supported by valid evidence if there exists at least one consistent instance:

$$\exists i \in \mathcal{I}(P_c) \;\; s.t. \; CV\left(a^{(i)}\right) \le \tau_{cv}, \tag{12}$$

where $\tau_{cv}$ is the consistency threshold. If the condition holds, we treat $P_c$ as providing suspicious structural evidence. For each clue node $c$, we maintain node sets $\mathcal{V}_{chains}(c)$ and $\mathcal{V}_{cycles}(c)$. Meanwhile, we maintain global structure sets $\mathcal{P}_{chain}$ and $\mathcal{P}_{cycle}$ to store all detected chain/cycle pattern instances:

$$\begin{aligned} &\mathcal{V}_{\text{type}}(c) \leftarrow \mathcal{V}_{\text{type}}(c) \cup \{v_0,\dots,v_L\}_{\text{type}}, \\ &\mathcal{P}_{\text{type}} \leftarrow \mathcal{P}_{\text{type}} \cup \{P_{\text{type}}\}, \;\; \text{type} \in \{chains, cycles\}. \end{aligned} \tag{13}$$

*Induced Subgraph*

Finally, SACC merges the candidate node sets and obtains the LIC node set for each clue $c$:

$$\mathcal{V}_{LIC}(c) = \{c\} \cup \mathcal{V}_{k-hop}(c) \cup \mathcal{V}_{chains}(c) \cup \mathcal{V}_{cycles}(c). \tag{14}$$

We construct $\mathcal{V}_{LIC}(c)$ for each $c \in C_V$ independently and take their union as the initial context node set $\mathcal{V}_0 = \bigcup_{c \in C_V} \mathcal{V}_{LIC}(c)$, which induces a candidate context subgraph $G[\mathcal{V}_0]$.

*Time Complexity*

The time complexity of SACC is (the full derivation is provided in supplementary material):

$$O\left(|C_V| \cdot d^k + |P| \cdot (m_{\max})^{L^*} \cdot L^*\right), \tag{15}$$

$|P| = \sum_{c \in C_V} |P_c|$, $L^* = \max(L_{max}, C_{max})$ and $d$ is the maximum node degree. $k, L^*$ and $m_{\max}$ are all small user-controlled constants, the overall cost remains confined to the clue-centered local neighborhood, which makes SACC practical for interactive investigation.

*B. Multi-Semantic Local-Temporal Graph Neural Network*

Given the local investigation context $\mathcal{G}_{LIC}$, our objective is to estimate a clue-conditioned risk field $\mathbf{p} \in [0,1]^{|\mathcal{V}_{LIC}|}$, where $p(v)$ measures the likelihood that account $v$ belongs to the laundering group implied by the current clues.

We identify two complementary evidence sources for account suspiciousness: (i) **event evidence**, reflecting coordinated transaction patterns within short time windows (e.g., structured splitting and rapid fund circulation) [24]; and (ii) **interaction evidence**, capturing how accounts collaborate through transactional relations and structural roles.

Accordingly, we propose MIST-GNN, which encodes these two evidence dimensions via a Local Temporal Context Encoder (LTCE) and a Multi-Semantic Relation Encoder (MSRE), respectively. Fig. 2 shows the overall architecture of MIST-GNN. Given $\mathcal{G}_{LIC}$, clue set $C_V$ and benign node set $B_V$, MIST-GNN outputs the conditional risk vector $\mathbf{p}$, forming a scored local investigation context $\mathcal{G}_{S-LIC} = (\mathcal{V}_{LIC}, \mathcal{E}_{LIC}, \mathbf{p})$.

*Account–Transaction Heterogeneous Graph*

To avoid mixing event attributes with multigraph edges and to explicitly separate account profiles from transaction behaviors, we node-ify transaction events and construct an account–transaction heterogeneous graph:

$$\mathcal{G}_{LIC}^{\mathcal{H}} = (\mathcal{V}_A \cup \mathcal{V}_T, \mathcal{E}_{\mathcal{H}}), \tag{16}$$

where $\mathcal{V}_A$ denotes account nodes and $\mathcal{V}_T$ denotes transaction nodes. For any transaction event $e$ with source account $u$ and destination account $v$, we add two relation edges $(u,e)$ and $(e,v)$. This construction disentangles account attributes from transaction-event features and provides the structural basis for multi-semantic channel design.

*Node Feature Encoding*

Account encoding. For an account node $v_a$ with raw feature vector $\mathbf{x}_a$, we use a two-layer MLP to obtain a static profile representation:

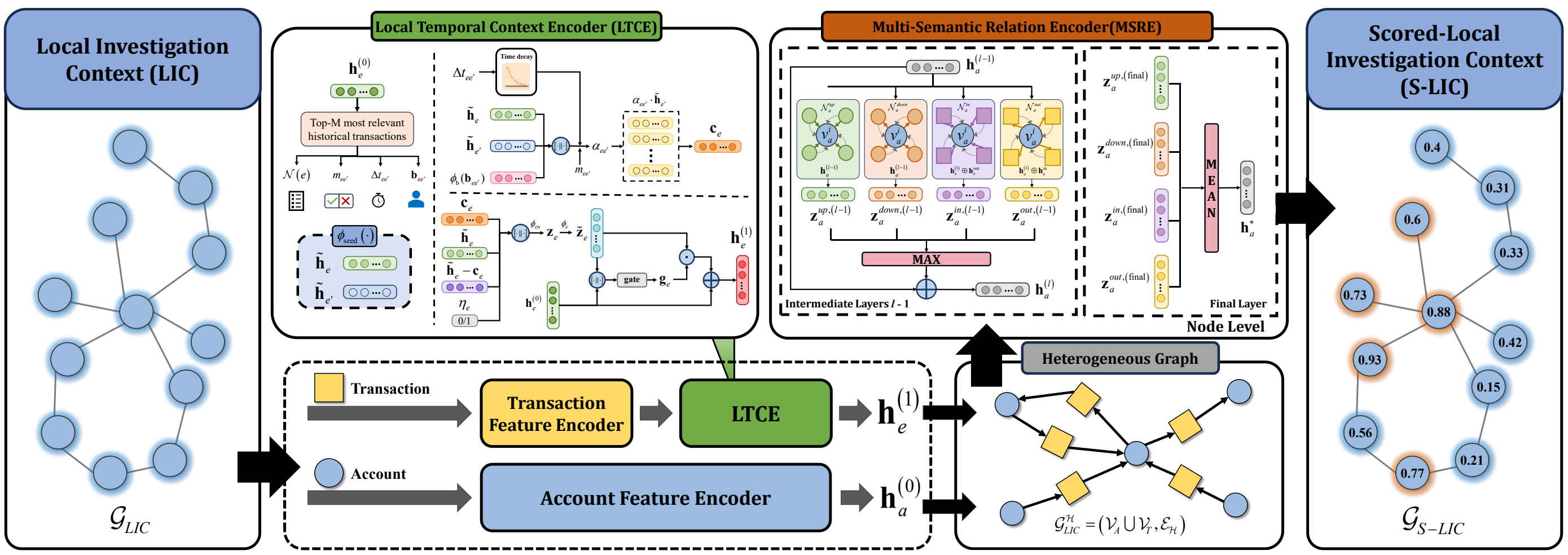

**Fig. 2.** Architecture of MIST-GNN, which consists of the LTCE and the MSRE and transforms the LIC into a Scored-LIC (S-LIC) carrying per-account risk scores.

$$\mathbf{h}_a^{(0)} = \mathrm{MLP}_A\left(\mathbf{x}_a\right), \tag{17}$$

Transaction encoding. A transaction node $e$ is associated with heterogeneous attributes. We encode each attribute type with an independent MLP channel, concatenate the resulting embeddings, and fuse them:

$$\mathbf{h}_e^{(0)} = \mathrm{MLP}_{\text{fusion}}\left(\Big\|_{s=1}^{S} \mathrm{MLP}_s\left(\mathbf{x}_e^{(s)}\right)\right), \tag{18}$$

*Local Temporal Context Encoder (LTCE)*

The representation of a transaction event depends not only on its own attributes but also on its surrounding context. LTCE aggregates this local context conditioned on role and time and injects the context signal into the transaction representation through explicit contrastive features and a gated residual.

For each event $e$, we preselect its $M$ most relevant historical transactions within the LIC and collect them into a neighbor index set $\mathcal{N}(e)$. To handle the case of fewer than $M$ neighbors, we attach a mask $m_{ee'} \in \{0,1\}$ to each neighbor $e' \in \mathcal{N}(e)$ indicating whether it is valid. For every neighbor $e'$, we build a role indicator vector $\mathbf{b}_{ee'}$ to describe its relation to $e$, and define the time gap $\Delta t_{ee'} = |t_e - t_{e'}|$ .(the details are provided in supplementary material)

To reduce the cost of context interactions, we first project $\mathbf{h}_e^{(0)}$ into a lower-dimensional context space, $\tilde{\mathbf{h}}_e = \phi_{\text{seed}}(\mathbf{h}_e^{(0)})$, and then compute attention scores. To encode the prior that more recent transactions are more relevant, we introduce a learnable decay factor $\delta = \mathrm{softplus}(\tilde{\delta})$ that acts on the time-gap term:

$$score_{ee'} = \phi_s\left([\tilde{\mathbf{h}}_e \| \tilde{\mathbf{h}}_{e'} \| \phi_b(\mathbf{b}_{ee'})]\right) - \delta \log\left(1 + \Delta t_{ee'}\right), \tag{19}$$

where $\phi_{\text{seed}}(\cdot), \phi_s(\cdot), \phi_b(\cdot)$ are shallow MLPs with non-linear activations. Logarithmic decay strikes a natural balance, sharp at short time scales and gentle at long ones, while avoiding numerical blow-up. Attention weights are computed via masked softmax with explicit re-normalization, giving the context representation $c_e$:

$$\alpha_{ee'} = \frac{\exp(score_{ee'})m_{ee'}}{\sum_{e' \in \mathcal{N}(e)} \exp(score_{ee'})m_{ee'} + \varepsilon}, \quad \mathbf{c}_e = \sum_{e' \in \mathcal{N}(e)} \alpha_{ee'} \cdot \tilde{\mathbf{h}}_{e'}. \tag{20}$$

We then concatenate the self-representation, the context, their difference, and a neighbor-validity flag $\eta_e = \mathbf{1}\{\mathcal{N}(e) \neq \varnothing\}$ to form an enhanced context representation:

$$\mathbf{z}_e = \phi_{\text{ctr}}\left([\tilde{\mathbf{h}}_e \| \mathbf{c}_e \| \tilde{\mathbf{h}}_e - \mathbf{c}_e \| \eta_e]\right). \tag{21}$$

The flag $\eta_e$ gives the downstream network an explicit signal in cold-start cases (no valid history). Finally, we lift $\mathbf{z}_e$ back to the base representation space and fuse it with the original representation via an adaptive gate:

$$\begin{aligned} \tilde{\mathbf{z}}_e &= \phi_e(\mathbf{z}_e), \qquad \mathbf{g}_e = \sigma\left(\mathbf{W}_g[\mathbf{h}_e^{(0)} \| \tilde{\mathbf{z}}_e]\right), \\ \mathbf{h}_e^{(1)} &= \mathrm{LN}\left(\mathbf{h}_e^{(0)} + \mathbf{g}_e \odot \tilde{\mathbf{z}}_e\right). \end{aligned} \tag{22}$$

The gate $\mathbf{g}_e$ lets the model rely on neighbors when the context is informative, and fall back to the self-representation when history is sparse or noisy, keeping the encoder robust across both dense and sparse entities. $\mathbf{h}_e^{(1)}$ is the final transaction representation produced by LTCE.

*Multi-Semantic Relation Encoder (MSRE)*

MSRE performs message passing over four semantic channels on the heterogeneous graph $\mathcal{G}_{LIC}^{\mathcal{H}}$, separately modeling "who transacts with whom" and "what transacts with what" patterns.

For an account node $a$, we define four semantic neighborhoods. The upstream and downstream counterpart channels capture account-level interaction:

$$\begin{aligned} \mathcal{N}_a^{up} &= \left\{u \,\middle|\, \exists e, (u,e) \in \mathcal{E}_{\mathcal{H}}, \wedge (e,a) \in \mathcal{E}_{\mathcal{H}}\right\}, \\ \mathcal{N}_a^{down} &= \left\{v \,\middle|\, \exists e, (a,e) \in \mathcal{E}_{\mathcal{H}}, \wedge (e,v) \in \mathcal{E}_{\mathcal{H}}\right\}. \end{aligned} \tag{23}$$

The outgoing and incoming event channels collect the transaction nodes directly connected to $a$ as sender or receiver, respectively.

$$\mathcal{N}_a^{out} = \left\{ e \middle| (a,e) \in \mathcal{E}_{\mathcal{H}} \right\},$$
$$\mathcal{N}_a^{in} = \left\{ e \middle| (e,a) \in \mathcal{E}_{\mathcal{H}} \right\}. \tag{24}$$

Let the account representation at layer $l$ be $\mathbf{h}_a^{(l)}$. For each channel $\kappa \in \{\text{up}, \text{down}, \text{out}, \text{in}\}$, we apply an independent GAT aggregator:

$$\mathbf{z}_a^{\kappa,(l)} = \text{GAT}^{\kappa,(l)}\left(\mathbf{h}_a^{(l)}, \left\{ \mathbf{h}_j \middle| j \in \mathcal{N}_a^{\kappa} \right\}\right), \tag{25}$$

where the neighbor representation $\mathbf{h}_j$ is defined according to the channel type. For the up and down channels, $\mathbf{h}_j$ is the account embedding. For the out and in channels, we use directional transaction representations that explicitly distinguish sending and receiving semantics, introducing role-aware contextual information beyond pure structural signals:

$$\mathbf{h}_j^{\text{out}} = \text{LN}(\mathbf{h}_e^{(1)} + \mathbf{h}_e^{\text{out}}),$$
$$\mathbf{h}_j^{\text{in}} = \text{LN}(\mathbf{h}_e^{(1)} + \mathbf{h}_e^{\text{in}}). \tag{26}$$

Where $\mathbf{h}_e^{\text{out}}$ represents the outgoing transaction features, capturing the historical spending patterns of the payer around the transaction, as well as the recent interaction intensity between accounts. $\mathbf{h}_u^{\text{in}}$ represents the incoming transaction features, which characterize the behavioral context when the transaction is treated as a fund inflow event.

To reduce redundancy across channels and stabilize training, we apply max pooling over the four channel outputs and update the account representation with a residual connection:

$$\mathbf{h}_a^{(l+1)} = \mathbf{h}_a^{(l)} + \max_{\kappa} \mathbf{z}_a^{\kappa,(l)}. \tag{27}$$

At the final layer, we retain all four channel outputs and project them into a shared latent space via channel-specific projections $\phi_\kappa$, then aggregate via mean pooling:

$$\mathbf{h}_a^{*} = \frac{1}{4}\sum_{\kappa} \phi_\kappa\left(\mathbf{z}_a^{\kappa,(\text{final})}\right). \tag{28}$$

The final membership probability is predicted as:

$$p_a = \sigma(\mathbf{W}^{\top}\mathbf{h}_a^{*} + b). \tag{29}$$

*Training Objective*

In the initial investigation stage, we adopt a PU learning setting, where clue nodes are treated as positives and all other nodes in the LIC are treated as unlabeled. Pseudo-negatives are selected from the unlabeled set based on the model's current prediction scores and their topological distance to the clues. To reduce label noise, we assign different weights to positive and pseudo-negative samples. Training consists of two stages: a warm-up stage with BCE loss on positives only, and a main stage with ranking loss and hard negatives. Details are given in the supplementary material. In the interactive investigation stage, analyst-confirmed benign nodes are introduced as explicit negatives, shifting the learning setting to PNU learning.

Accordingly, the model is trained with a supervised binary cross-entropy loss, combined with ranking constraints and cross-semantic consistency regularization.

For labeled indices, we use binary cross-entropy:

$$\mathcal{L}_{\text{BCE}} = -\sum_i \left[ y_i \log p_i + (1-y_i)\log(1-p_i) \right]. \tag{30}$$

A margin-based ranking regularizer encourages separation between positive and comparison nodes:

$$\mathcal{L}_{rank} = \mathbb{E}\left[ m - (\text{logit}_i - \text{logit}_j) \right]_+, \tag{31}$$

where $\text{logit}_i$ and $\text{logit}_j$ denote the predicted logits of a positive node and a comparison node, respectively, $m$ is the margin, and $[x]_+ = \max(0, x)$.

To encourage consistent risk patterns across semantic channels, we introduce a consistency regularization in the shared latent space:

$$\mathcal{L}_{con} = \sum_{\kappa} \left\| \phi_\kappa\left(\mathbf{z}_a^{\kappa,(\text{final})}\right) - \mathbf{h}_a^{*} \right\|^2. \tag{32}$$

The final loss is:

$$\mathcal{L} = \mathcal{L}_{\text{BCE}} + \lambda_1 \mathcal{L}_{rank} + \lambda_2 \mathcal{L}_{con}. \tag{33}$$

During inference, the model outputs node risk scores $\mathbf{p}$ and yield a scored-local investigation context $\mathcal{G}_{S-LIC}$ for the downstream assembly module.

*C. Evidence-Driven Group Assembly*

After obtaining the scored local investigation context $\mathcal{G}_{S-LIC} = (\mathcal{V}_{LIC}, \mathcal{E}_{LIC}, \mathbf{p})$, EDGA recovers the member set $Y^{(r)*}$ of the target laundering group starting from clue component $C_V^{(r)}$. The group is initialized with the clue nodes and grown iteratively by selecting the most evidentially supported boundary candidate at each step. The process terminates when stopping criteria are met, yielding an induced subgraph $\mathcal{G}_{S-LIC}\left[Y^{(r)*}\right]$ as the recovered group associated with the *r*-th clue component. When the clue set contains multiple connected components, EDGA produces a set of recovered group views, one for each component. They provide different clue-conditioned perspectives on the underlying laundering group and can support subsequent investigation.

*Greedy Expansion*

We initialize $Y_0 = C_V^{(r)}$ and restrict expansion to the 1-hop boundary:

$$\mathcal{B}(Y_t) = \left\{ u \notin Y_t \middle| \exists v \in Y_t, (v,u) \in \mathcal{E}_{LIC} \right\}. \tag{34}$$

At each step we add the candidate with the highest marginal gain:

$$u^{*} = \arg\max_{u \in \mathcal{B}(Y_t)} \Delta J\left(u \middle| Y_t\right), \quad Y_{t+1} \leftarrow Y_t \cup \{u^{*}\}. \tag{35}$$

This repeats until a stopping criterion is met.

*Evidence Signals*

We construct three complementary signals to measure the marginal benefit of adding a candidate node $u$ to the current group $Y$.

**(1) Risk Consistency.** MIST-GNN outputs a node-level risk field $\mathbf{p}$, where $p(u)$ represents the likelihood that account $u$ belongs to the target laundering group. We define the risk-consistency signal as:

$$s_R(u) = p(u). \tag{36}$$

This signal is particularly informative in the early stages of expansion, driving the group to concentrate quickly on high-risk regions.

**(2) Structural Consistency.** Structure entropy [30] quantifies the organizational quality of a graph by measuring the average coding length of a random walk under a given partition. A lower value indicates denser intra-cluster connections and sparser boundary cuts. To capture such structural coherence in transaction networks, we adopt the two-dimensional structure entropy, which restricts the partition tree height to 2:

$$\mathcal{H}^2(G) = \min_{T:height(T)=2} \mathcal{H}^T(G). \tag{37}$$

During assembly, we treat the current group $Y$ as one cluster and all remaining nodes as singletons. The structural-consistency signal is the entropy reduction from adding $u$:

$$s_E(u) = \mathcal{H}^2\left(\mathcal{G}_{S-LIC}[Y]\right) - \mathcal{H}^2\left(\mathcal{G}_{S-LIC}\left[Y \cup \{u\}\right]\right). \tag{38}$$

To enable efficient computation, we adopt a local incremental formulation [31] , which updates the entropy using only local quantities without recomputing it over the full graph. Let $w(Y,u)$ denote the total connection weight between $u$ and $Y$ , $vol(Y)$ the volume of $Y$ , and $vol\left(\mathcal{G}_{S-LIC}[Y]\right)$ the internal edge volume of $\mathcal{G}_{S-LIC}[Y]$. Then:

$$s_E(u) = f\left(w(Y,u), vol(Y), vol\left(\mathcal{G}_{S-LIC}[Y]\right)\right), \tag{39}$$

where $vol\left(Y \cup \{u\}\right)$ and $vol\left(\mathcal{G}_{S-LIC}\left[Y \cup \{u\}\right]\right)$ are updated incrementally from these local quantities. Thus, at each step we only need to maintain and update these three local quantities, making the computation $O(d_u)$ per iteration.

**(3) Prior Structure.** SACC enumerates chain and cycle instances that pass amount consistency checks during context construction. We reward candidates that advance the completeness and continuity of these patterns. For a pattern instance $P$ with node set $V_P$ , the coverage count given current group $Y$ is $Q(P,Y) = |Y \cap V_P|$.

Cycles reward late-stage closure, so we use a convex completion function:

$$\begin{aligned} f_{cy}(Q) &= \left(\frac{Q}{|P|}\right)^2, \\ \delta_{cy}(u,P) &= f_{cy}\left(Q\left(P, Y \cup \{u\}\right)\right) - f_{cy}\left(Q(P,Y)\right). \end{aligned} \tag{40}$$

Chains reward early path formation and diminish as the chain extends, so we use a concave function:

$$\begin{aligned} f_{ch}(Q) &= \sqrt{\frac{Q}{|P|}}, \\ \delta_{ch}(u,P) &= f_{ch}\left(Q\left(P, Y \cup \{u\}\right)\right) - f_{ch}\left(Q(P,Y)\right). \end{aligned} \tag{41}$$

Beyond node completeness, we introduce a continuous coverage reward based on adjacent-edge coverage. Let $\mathrm{con}(Y,P)$ the number of adjacent edges covered by $Y$ .

$$\delta_{edge}(u,P) = \mathrm{con}\left(Y \cup \{u\}, P\right) - \mathrm{con}(Y,P). \tag{42}$$

The combined prior-structure gain for cycles and chains is:

$$\begin{aligned} s_P^{cy}(u) &= \max_P\left(\delta_{cy}(u,P) + \delta_{edge}(u,P)\right), \\ s_P^{ch}(u) &= \max_P\left(\delta_{ch}(u,P) + \delta_{edge}(u,P)\right). \end{aligned} \tag{43}$$

Using max over instances avoids double counting when a node participates in multiple patterns, and it makes each candidate primarily explained by its strongest structural evidence. The final prior-structure signal is:

$$s_P(u) = s_P^{cy}(u) + s_P^{ch}(u). \tag{44}$$

*Rank-based Three-evidence Voting*

Considering the heterogeneity of the three signals, we adopt a rank-based three-evidence voting mechanism. For each signal, candidates in the current boundary are ranked in descending order according to their scores. The rank $r(u)$ is then linearly mapped to a strength score:

$$g(u) = \frac{r_{max} - r(u)}{r_{max} - 1} \in [0,1]. \tag{45}$$

For the risk signal $s_R$ and the structural entropy signal $s_E$ , we select the top $\rho_R$ and top $\rho_E$ proportions of boundary candidates for voting. Since the prior signal $s_P$ is sparse, a candidate receives one vote from this signal as long as $s_P(u) > 0$ . The three binary votes are defined as:

$$\begin{aligned} v_R(u) &= \mathbf{1}\left[\frac{r_R(u)}{|\mathcal{B}|} \le \rho_R\right], v_E(u) = \mathbf{1}\left[\frac{r_E(u)}{|\mathcal{B}|} \le \rho_E\right], \\ v_P(u) &= \mathbf{1}\left[s_P(u) > 0\right]. \end{aligned} \tag{46}$$

Let $V(u) = v_R(u) + v_E(u) + v_P(u)$ , and the multi-evidence candidate set is defined as:

$$\mathcal{B}^+ = \left\{u \in \mathcal{B}(Y) \middle| V(u) \ge 2\right\}, \tag{47}$$

a candidate is retained only if it is supported by at least two evidence sources. For candidates in $\mathcal{B}^+$ , we compute the final fusion score as:

$$\Delta J(u|Y) = \lambda_E g_E(u) + \lambda_R g_R(u) + \lambda_P g_P(u) + 0.10\frac{V(u)}{3}. \tag{48}$$

These weights are constrained by $\lambda_E + \lambda_R + \lambda_P = 0.9$ , so that the remaining 0.10 is reserved for the voting-consistency reward $V(u)/3$ , which encourages candidates supported by multiple sources. If multiple candidates have the same score, ties are broken in the order of $V(u)$, structural consistency, risk consistency, and prior consistency.

*Stopping Criterion*

As $Y$ grows, its boundary may become increasingly loose. Therefore, we use the conductance $\mathrm{cond}(Y)$ of the current set to monitor the expansion quality. We define the improvement at step $t$ as $\Delta_t = \log\left(\mathrm{cond}(Y_{t-1})\right) - \log\left(\mathrm{cond}(Y_t)\right)$ . If $\Delta_t < \varepsilon_{cond}$ for two consecutive steps, we regard the expansion as entering a plateau stage. After entering this stage, the algorithm continues expanding only when there exists a strong-evidence candidate, namely a candidate satisfying either $s_P(u) > \tau_P$ or $s_R(u) > \tau_R$ . Otherwise, the algorithm stops. In addition, the expansion also terminates when $|Y| \geq K_{max}$ or when the boundary is exhausted.

## V. Experiments

This chapter presents the experimental evaluation of the Clue2Group framework. We organize the experiments in a hierarchical manner and analyze the framework through four research questions.

Q1: Given sparse initial clues and different clue positions, can SACC construct high-quality local investigation contexts (LICs)?

Q2: Under PU supervision in the initial stage, can MIST-GNN produce a stable and effective conditional risk field?

Q3: Can EDGA recover money laundering group boundaries more accurately? Meanwhile, is the evidence-driven group assembly mechanism stable and robust?

Q4: In the interactive setting, can Clue2Group outperform the general ICS-GNN baseline? Is the framework robust to different clue selections, computationally efficient for interactive investigation, and able to provide interpretable evidence chains through case studies?

**Datasets.** We evaluate our framework on the public AML benchmark AMLSim [32], a widely-adopted simulation benchmark encompassing 8 money laundering typologies. We adopt two configurations of varying scale: **HI-Small**, which contains approximately 500K accounts and 5M transactions, and **HI-Medium**, which further scales up to ~2M accounts and 32M transactions, allowing us to assess the scalability and robustness of our method. Notably, the dataset provides malicious labels only at the transaction level. We construct the ground-truth groups using a unified strategy: we first induce a subgraph containing all accounts involved in illicit transactions, and then identify each connected component within that subgraph as a distinct money laundering group.

### *Evaluation on SACC (Q1)*

**Implementation Details.** To evaluate the ability of SACC to construct LICs under different clue positions and hop settings, we randomly selected 20 groups for evaluation. For each group, we divided the accounts in its induced subgraph into core nodes and peripheral nodes according to node degree, and then randomly selected five clue nodes from each category. In each experiment, we used a single clue node as the starting point and compared four LIC construction methods under $k \in \{1,2,3\}$ : neighborhood expansion (NE), NE+ch with chain-structure, NE+cy with cycle-structure, and SACC. We report the average Recall in Table 1, as the primary goal of LIC construction is to cover as many true group members as possible. Precision and F1 are provided in supplementary material. SACC also achieves strong performance on these metrics, indicating that the improvement does not come from blindly expanding the local context.

From Table I, SACC shows consistent advantages across different clue positions, k-hop settings, and both datasets. This indicates that the LIC constructed by SACC is reliable in covering true group members. As $k$ increases, all methods improve noticeably, suggesting that enlarging the local search scope helps include more true members. However, under the same $k$, SACC consistently outperforms or matches the baselines. This shows that its gains come not only from a larger neighborhood, but also from the additional evidence introduced by structural enhancement.

TABLE I

Recall of LIC construction methods under different clue positions and k-hop.

| k-hop | method | HI-Small | | HI-Medium | |
|---|---|---|---|---|---|
| | | Core | Peripheral | Core | Peripheral |
| k=1 | NE | 0.3333 | 0.1887 | 0.2872 | 0.1670 |
| | NE+ch | 0.4590 | 0.3006 | 0.4329 | 0.2853 |
| | NE+cy | 0.3610 | 0.2148 | 0.2872 | 0.1670 |
| | SACC | **0.4702** | **0.3118** | **0.4329** | **0.2853** |
| k=2 | NE | 0.7935 | 0.7786 | 0.6554 | 0.5956 |
| | NE+ch | 0.8643 | 0.8388 | 0.7434 | 0.6663 |
| | NE+cy | 0.8121 | 0.7973 | 0.6554 | 0.5956 |
| | SACC | **0.8718** | **0.8463** | **0.7434** | **0.6663** |
| k=3 | NE | 0.8742 | 0.8546 | 0.7824 | 0.7175 |
| | NE+ch | 0.9170 | 0.8924 | 0.8362 | 0.7620 |
| | NE+cy | 0.8853 | 0.8658 | 0.7824 | 0.7175 |
| | SACC | **0.9207** | **0.8962** | **0.8362** | **0.7620** |

The advantage is most pronounced at $k = 1$ . This suggests that when the initial clue provides limited context, chain and cycle augmentation can effectively mitigate structural fragmentation in the LIC. In addition, results from core clues are generally better than those from peripheral clues, indicating that the clue position affects how well the local context can be recovered. Still, SACC maintains clear improvements in peripheral settings, demonstrating stronger robustness under weak clues.

It is also worth noting that on HI-Medium, SACC achieves the same performance as NE+ch. This suggests that the main gain in this dataset comes from chain-based enhancement, while the additional benefit from cycle structures is relatively limited.

### *Evaluation on MIST-GNN (Q2)*

**Implementation Details.** We randomly select 20 laundering groups (each with ≥10 accounts) and repeat each experiment 3 times. In each trial, we simulate the initial investigation stage, where the clue set contains only one connected component. Specifically, ~10% of intra-group transactions are sampled via connected expansion and the involved accounts serve as clue nodes. LICs are constructed by SACC with $k = 2$ , and their size is constrained to [100, 15000] nodes to exclude abnormally small or large subgraphs. Training follows a PU setting: clue accounts are positive, remaining LIC nodes are unlabeled, with pseudo-negatives mined dynamically. For evaluation, we treat only the accounts that belong to the target laundering group and are contained in the LIC, excluding clue nodes, as ground-truth positives. Performance is measured by AUC and AP.

**Baselines and Variants.** We compare against six GNN baselines spanning homogeneous (GCN, GAT, GraphSAGE) and heterogeneous (R-GCN, HAN, HGT) paradigms. Ablation studies remove or replace key components (w/o LTCE, ac-channel only, tx-channel only, 1-channel, w/o tx(out & in)). Variant studies substitute design choices in fusion (mean/attention/MLP), convolution (GCN layer). All baselines

TABLE II

PERFORMANCE OF MIST-GNN AND COMPARED MODELS UNDER SPARSE PU SUPERVISION.

| Type | Model | HI-Small | | HI-Medium | |
|---|---|---|---|---|---|
| | | AUC | AP | AUC | AP |
| | GCN | 0.4958 | 0.1125 | 0.4689 | 0.1519 |
| Homo | GAT | 0.6884 | 0.1576 | 0.5933 | 0.1682 |
| | GraphSAGE | 0.7437 | 0.2322 | 0.6900 | 0.2932 |
| | R-GCN | 0.7529 | 0.1556 | 0.6184 | 0.1898 |
| Hetero | HAN | 0.5309 | 0.0988 | 0.4607 | 0.1286 |
| | HGT | 0.4894 | 0.0708 | 0.4818 | 0.1206 |
| | w/o LTCE | 0.7645 | 0.2686 | 0.7611 | 0.3035 |
| | ac-channel | 0.6552 | 0.2151 | 0.6675 | 0.2192 |
| Ablation | tx-channel | 0.6880 | 0.1532 | 0.6077 | 0.1630 |
| | 1-channel | 0.7123 | 0.2017 | 0.6416 | 0.1735 |
| | w/o tx(out & in) | 0.7460 | 0.2674 | 0.6991 | 0.2303 |
| | mean fusion | 0.7682 | 0.2658 | 0.7508 | 0.3018 |
| Variant | attention fusion | 0.7255 | 0.2109 | 0.7216 | 0.2543 |
| | mlp fusion | 0.7633 | 0.2340 | 0.7234 | 0.2853 |
| | GCNlayer | 0.7766 | 0.2562 | **0.7699** | 0.2827 |
| | MIST-GNN | **0.7879** | **0.2945** | 0.7569 | **0.3048** |

are tuned; all MIST-GNN variants share identical hyperparameters.

Q2 corresponds to a challenging scenario at the early stage of an investigation. During training, only a small number of clue accounts are treated as positive samples, while all remaining LIC nodes are unlabeled. Therefore, the current results can be viewed as a conservative estimate of model performance in the initial investigation stage. In practical use, as investigators gradually confirm additional positive or negative samples, this feedback can be injected back into MIST-GNN to further improve the quality of the risk field.

As shown in Table II, MIST-GNN achieves the best AUC and AP on HI-Small, and the best AP on HI-Medium. Compared with homogeneous GNNs and standard heterogeneous GNNs, its advantage mainly comes from its multi-semantic modeling design for money laundering scenarios. Homogeneous models tend to ignore the distinction between accounts and transaction events. Standard heterogeneous models also do not explicitly separate fund inflows, fund outflows, and upstream or downstream account relations. By decomposing the neighborhood into upstream accounts, downstream accounts, outgoing transactions, and incoming transactions, MIST-GNN captures both account-level interaction patterns and event-level transaction behavior.

The results further show that account-based and transaction-based channels provide complementary information. When only a single semantic channel is used, the performance is consistently lower than that of the full model. Removing directional transaction features also leads to a clear performance drop, suggesting that distinguishing between fund inflows and outflows is important for modeling laundering behavior. The contribution of LTCE depends on whether the temporal scale matches the data. When the time window and neighbor sampling strategy can properly cover the true transaction rhythm of target groups, as in HI-Small, LTCE brings a clear improvement, increasing AUC from 0.7645 to 0.7879. In contrast, on HI-Medium, where transactions are sparser and span a longer time range, its gain is more limited. This suggests that temporal context modeling should be adjusted according to the transaction density of different datasets. We further discuss this issue in the supplementary material.

Overall, MIST-GNN provides a domain-aware risk ranking signal under sparse PU supervision. As more human feedback becomes available during the interactive investigation process, the risk field can be further refined. This signal is also complementary to the structural entropy and chain/cycle prior evidence used later in EDGA.

*Evaluation on EDGA (Q3)*

**Implementation Details.** Group sampling, LIC construction, and risk-field generation follow the same settings as in Q2. For EDGA, we set $\rho_R = 0.2$ and $\rho_R = 0.5$, corresponding to the top 20% and top 50% of boundary candidates. In the plateau stage, we set $\tau_R = 0.95$ for the risk signal and $\tau_R = 0.8$ for the prior signal. The weights $(\lambda_R, \lambda_E, \lambda_P)$ are set to the default configuration (0.45, 0.35, 0.10). The maximum expansion budget is set to $K_{max} = 80$. For evaluation, we treat only the accounts that belong to the target laundering group and are contained in the LIC, excluding clue nodes, as ground-truth positives. We report the average Precision, Recall, and F1 score over all cases.

**Baselines.** We compare against two groups of assembly baselines: general community detection methods, including Louvain and LPA, and random-walk-based local expansion methods, including PPR-Nibble and Weighted-PPR. Weighted-PPR incorporates the MIST-GNN risk field as node weights. All baselines are evaluated on the same LICs and clue nodes, and all risk fields are generated by MIST-GNN.

**Analysis**. Table III compares EDGA with different baseline methods on the group recovery task. EDGA achieves the highest Precision and F1 on both HI-Small and HI-Medium, showing better accuracy and overall recovery quality. Although PPR-Nibble and WPPR-Nibble obtain higher Recall on both datasets, their Precision is clearly lower than that of EDGA. This suggests that these random-walk-based methods tend to cover more true members through more aggressive boundary expansion, but also introduce many background nodes. Table IV further explains this result from the perspective of recovered group size. Compared with the baselines, EDGA recovers a group size that is closer to the ground-truth group size within the LIC. This is especially clear on HI-Medium, where the average ground-truth group size is 13 and EDGA recovers 12 nodes on average. Overall, the three-signal design of EDGA provides complementary constraints during expansion. It helps reduce the inclusion of noisy nodes while preserving valid group members, leading to a better Precision–Recall trade-off.

*Sensitivity Analysis of EDGA Fusion Weights*

**Implementation Details.** We evaluate the sensitivity of EDGA to the fusion weights of the three evidence signals $(\lambda_R, \lambda_E, \lambda_P)$ on a ternary simplex. In total, we test 22 configurations. 3 vertex points for single-signal ablations, 12 edge points covering four mixing ratios for each pair of signals, 6 interior points for three-signal fusion, and the default configuration.

TABLE III
PERFORMANCE COMPARISON OF EDGA AND BASELINE METHODS.

| Method | HI-Small | | | HI-Medium | | |
|---|---|---|---|---|---|---|
| | Prec. | Rec. | F1 | Prec. | Rec. | F1 |
| Louvain | 0.112 | 0.312 | 0.144 | 0.160 | 0.316 | 0.197 |
| LPA | 0.226 | 0.236 | 0.163 | 0.258 | 0.201 | 0.210 |
| PPR-Nibble | 0.103 | 0.465 | 0.163 | 0.156 | 0.601 | 0.234 |
| WPPR-Nibble | 0.223 | **0.607** | 0.260 | 0.209 | **0.613** | 0.265 |
| EDGA | **0.303** | 0.325 | **0.273** | **0.351** | 0.312 | **0.286** |

TABLE IV
SIZE COMPARISON OF LICS, GROUND-TRUTH GROUPS, AND RECOVERED GROUPS

| | HI-Small | HI-Medium |
|---|---|---|
| **GT group size in LIC** | 16 | 13 |
| **LIC subgraph size** | 2298 | 1089 |
| Louvain | 82 | 25 |
| LPA | 368 | 9 |
| PPR-Nibble | 69 | 43 |
| WPPR-Nibble | 230 | 40 |
| EDGA | **37** | **12** |

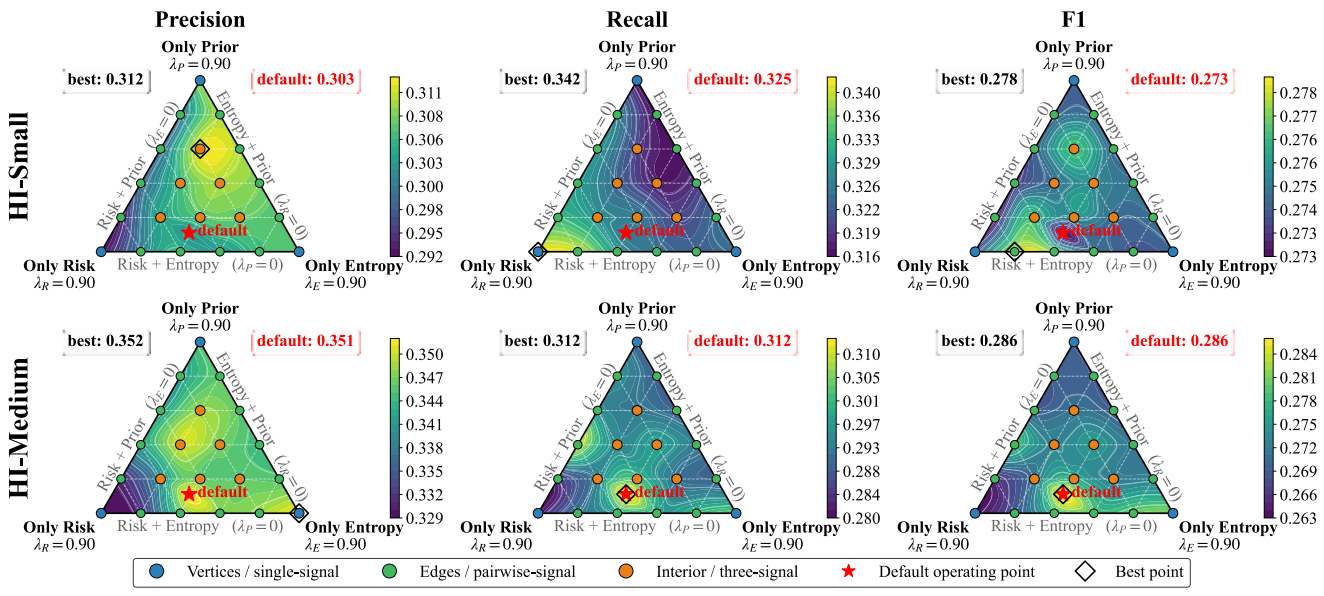

**Fig.3.** Ternary simplex analysis of EDGA fusion weight sensitivity.

**Analysis.** Fig. 3 reports Precision, Recall, and F1 on the two datasets. The results show that the fusion mechanism is robust to weight selection. The performance variation is small across most panels. For example, on HI-Small, F1 only varies within [0.273, 0.278]. The default configuration is within 1.8% of the best F1. On HI-Medium, the default configuration achieves the best Recall and F1, while its Precision is only 0.2% lower than the best value. These results indicate that EDGA does not rely on dataset-specific weight tuning. The three evidence signals play complementary roles. On HI-Small, the risk signal achieves the highest Recall but relatively lower Precision, suggesting that it tends to expand the group more aggressively. In contrast, structural entropy and prior pattern evidence provide stronger constraints on the assembly process and lead to more conservative outputs. The best-performing regions also differ across datasets, showing that no single signal consistently dominates all metrics. Overall, three-signal fusion provides a more balanced performance. In particular, on HI-Medium, the default configuration achieves an F1 of 0.286, outperforming the best single-signal and two-signal configurations. On HI-Small, although one risk–entropy configuration achieves a slightly higher F1 than the default setting, the default still remains close to the optimum and provides a more stable Precision–Recall trade-off.

In summary, the three evidence signals in EDGA are complementary. The fusion mechanism remains stable across a wide range of weight configurations, and the default weights lie within this robust region. Therefore, EDGA does not require fine-grained weight tuning.

*Evaluation on Clue2Group (Q4)*

**Implementation Details.** Group sampling follows the same setting as in Q2. In each experiment, the initial clue set contains only one connected component, which is obtained by sampling approximately 10% of the intra-group transactions. Each experiment runs for 10 interaction rounds. After the target group is produced in each round, if the output contains true group members, we randomly select one of them and add it to the clue set as a positive label. If the output contains benign accounts, we randomly select one of them and add it to the benign node set as a negative label. If no such node exists, the label set is not updated in that round. For evaluation in each interaction round, we exclude the clue nodes and report Precision, Recall, and F1.

**Baselines.** We use **ICS-GNN**[14] as the main baseline for two reasons. First, ICS-GNN and Clue2Group follow a similar task paradigm. Both are interactive, feedback-driven frameworks for query-conditioned subgraph discovery. Second, ICS-GNN is designed for general social network scenarios, whereas Clue2Group is a domain-specific framework for money laundering investigations. Comparing them directly allows us to examine whether domain-specific design is necessary beyond a general interactive framework. To avoid letting the comparison be dominated by differences in representation learning capacity, we further introduce a variant named **ICS-MIST-GNN**. In this variant, the GCN module in ICS-GNN is replaced with MIST-GNN. This variant is used to test whether simply upgrading the GNN backbone can close the performance gap between a general framework and a domain-specific framework.

For Clue2Group, the LIC is constructed by SACC with $k$=1 and without a size constraint. This setting simulates a realistic scenario where the LIC is not required to cover all members of the laundering group. The hyperparameters of EDGA follow the Q3 setting. Since ICS-GNN and its variant require the output group size to be specified in advance, while Clue2Group does not impose such a constraint, we set the output size of both baselines in each round to be the same as the group size produced by EDGA in that round for a fair comparison.

**Analysis**. Fig. 4 shows the performance of the three methods over 10 interaction rounds on two datasets. The evaluation excludes clue nodes and therefore measures the ability to identify new group members. Clue2Group consistently outperforms ICS-GNN and ICS-MIST-GNN on Precision, Recall, and F1. This supports two assumptions: domain-specific design is more effective than general interactive community search in AML investigations, and the advantage of Clue2Group cannot be achieved by simply replacing the GNN backbone, as ICS-MIST-GNN still underperforms Clue2Group.

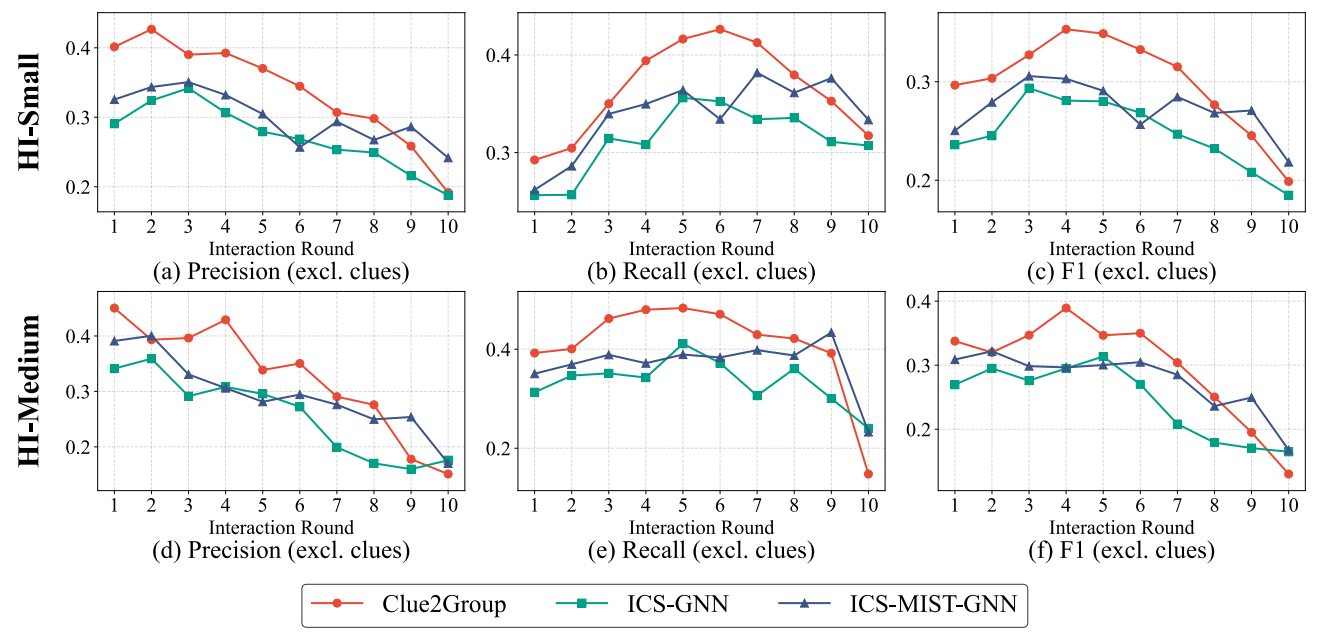

**Fig.4.** Interactive group recovery performance under excluding-clue evaluation.

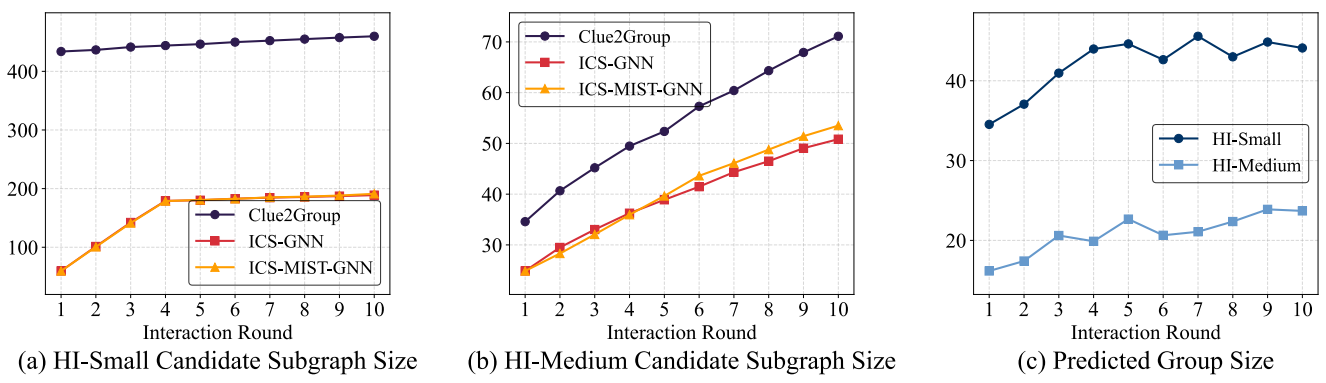

**Fig.5.** Candidate subgraph size and output size over interaction rounds.

The decline in Precision is mainly caused by the growing candidate space. As shown in Fig. 5, both the candidates subgraph size and the predicted group size increase over interaction rounds. The model therefore needs to make decisions in a larger and noisier search space, making it harder to maintain high precision. Recall and F1 first increase and then decrease on both datasets, with their peaks appearing around rounds 4–6. The early increase shows that interactive feedback helps expand the group boundary and recover more true members. The later decline indicates that the benefit of additional feedback is gradually offset by precision loss. This suggests that interactive AML investigation has an optimal interaction window, rather than improving monotonically with more rounds. This is also consistent with practical AML settings, where analyst feedback is costly and timely decisions are required. Results including clue nodes are reported in the supplementary material to show the cumulative discovery effect perceived by analysts.

It is also worth noting that Clue2Group shows a larger drop in later rounds than the baselines. This is because its multi-evidence fusion relies on the quality of risk, structural, and prior-pattern signals. As the candidate boundary moves farther from the original clues, these signals become weaker. Structural entropy becomes less informative, chain/cycle priors provide fewer constraints, and MIST-GNN may accumulate bias under limited supervision. This suggests that EDGA should be paired with a proper stopping strategy to avoid expansion when evidence becomes sparse.

*1) Clue Sensitivity Analysis*

*Clue Quantity Sensitivity*

**Implementation Details.** To evaluate the robustness of Clue2Group to the size of the initial clue set, we conduct experiments on HI-Small by randomly sampling 3%, 5%, 10%, and 20% of nodes from each true laundering group as initial clues.

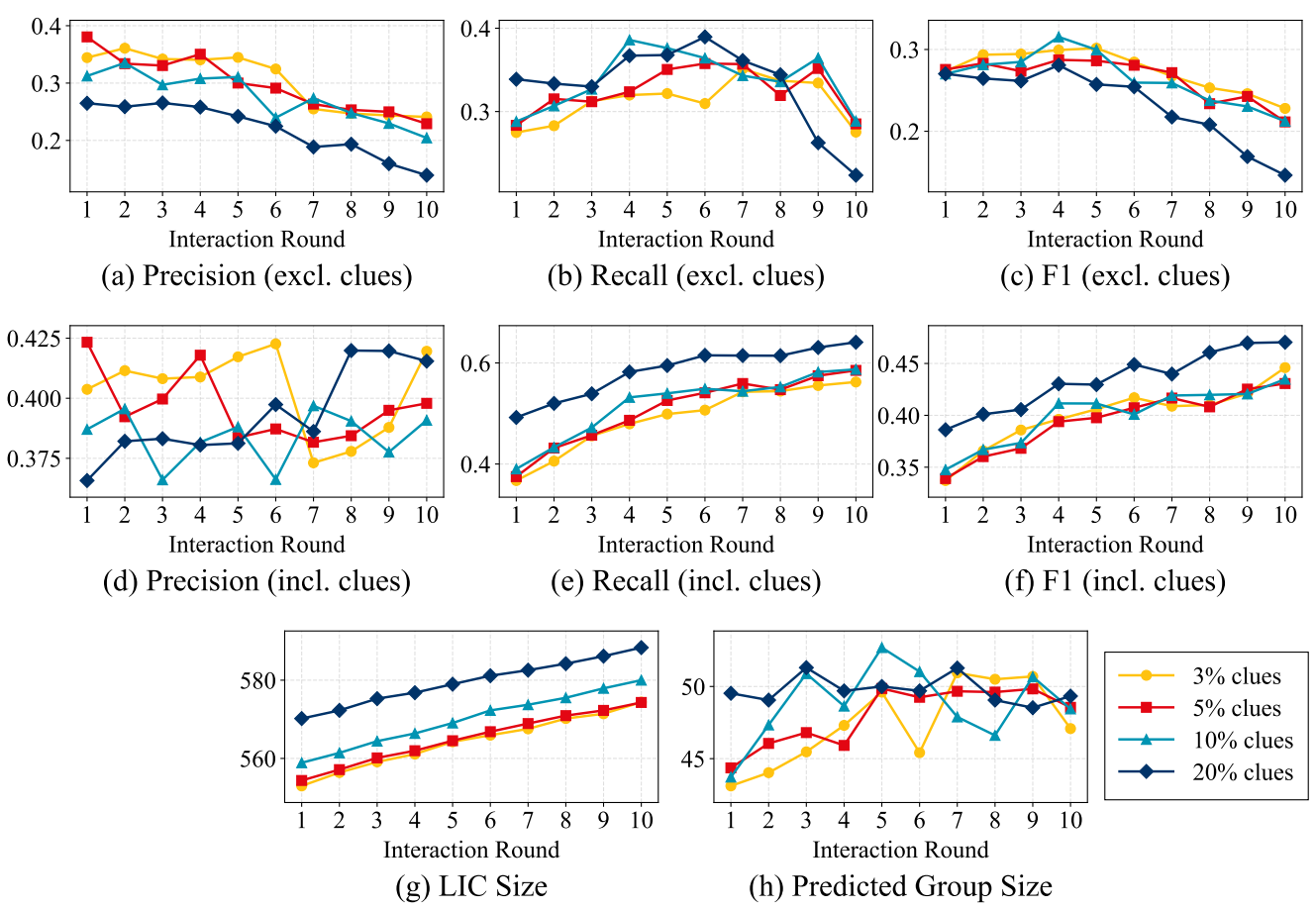

**Fig.6.** sensitivity to initial clue quantity.

**Analysis**. Fig. 6 (a)–(c) shows the results when clue nodes are excluded from evaluation. Under the four initial clue ratios, the Precision, Recall, and F1 curves show largely consistent trends. The gaps in Recall and F1 are also small in the early and middle rounds, indicating that Clue2Group is stable under different initial clue sizes. Fig. 6(h) further shows that the predicted group sizes gradually converge across different clue ratios. The cross-setting range decreases from about 6.5 in round 1 to about 1.8 in round 10. This suggests that the final output size does not grow linearly with the initial clue ratio or the LIC size.

The 20% clue setting shows relatively lower Precision and F1 in most rounds. This is mainly due to the evaluation protocol that excludes clue nodes. On the one hand, a larger initial clue set removes more true group members from the evaluation set, leaving a smaller and harder residual target set. On the other hand, as shown in Fig. 6(g), the LIC constructed under the 20% clue setting is consistently larger, introducing more irrelevant nodes. Meanwhile, Fig. 6(h) shows that the predicted group sizes under different settings gradually converge. Therefore, the model needs to select a similar number of nodes from a larger search space, which increases the chance of false positives.

Fig. 6(d)–(f) reports the results when clue nodes are included in evaluation. This setting is closer to real investigation scenarios, where analysts care about the final group output formed by both initial clues and model-expanded nodes. From this perspective, the 20% clue setting achieves clearly better Recall and F1. This indicates that more initial clues provide richer structural information and stronger supervision, thereby improving the overall quality of group recovery.

Overall, Clue2Group maintains stable discovery ability under sparse clue settings, while further improving the practical investigation output when more clues are available.

*Clue Position Sensitivity*

**Implementation Details**. To evaluate the robustness of Clue2Group to the position of the initial clues, we conduct experiments on the HI-Small and construct two types of clue sets based on account node degrees in each true group-induced subgraph. Specifically, we select the top 10% highest-degree

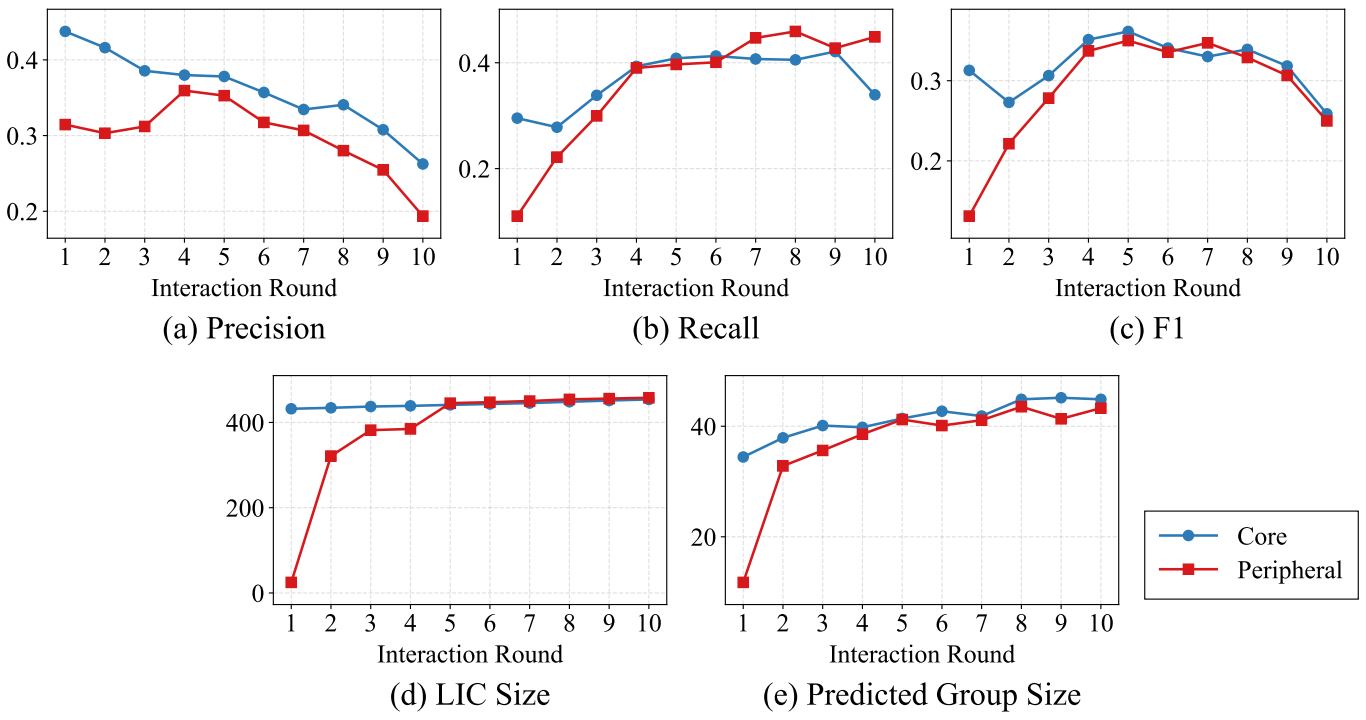

**Fig.7.** sensitivity to initial clue position.

nodes as core clues and the bottom 10% lowest-degree nodes as peripheral clues.

**Analysis**. Fig. 7 reports the sensitivity analysis with respect to clue position. Overall, core clues show a clear advantage in the early interaction rounds. Since core nodes are located in denser and more structurally representative regions of the target group, they help the framework construct a more complete LIC from the first round. As shown in Fig. 7(d), the LIC size under core clues in the first round is nearly 20 times larger than that under peripheral clues. This also leads to larger initial predictions and better early-stage performance, as shown in Fig. 7(e), Fig. 7(b), and Fig. 7(c).

However, the performance under peripheral clues improves rapidly as the interaction proceeds. As analyst-validated true members are added to the clue set in each round, both the LIC size and the predicted group size expand quickly. Around round 5, they become close to those under core clues. The corresponding Recall and F1 also converge, with F1 intersecting around round 4. This shows that the interactive mechanism of Clue2Group can effectively reduce the negative impact of unfavorable initial clue positions.

Overall, clue position mainly affects the early localization efficiency of the framework. Core clues lead to higher-quality initial predictions, while peripheral clues start from a weaker position but can gradually approach the recovery quality of core clues through multi-round interaction. This indicates that Clue2Group does not fully depend on high-quality initial clues. Instead, it can use interactive feedback to compensate for the limited information provided by weak clues and eventually reach similar steady-state performance.

*2) Efficiency Analysis*

**Implementation Details**. To evaluate the runtime efficiency of Clue2Group, we conduct the experiment on the HI-Small dataset. We record the per-round runtime of SACC, MIST-GNN, EDGA, and the full pipeline over 10 interaction rounds. We then compute the average per-round runtime of each module to measure the overall computational cost of the framework during interactive investigation.

**Analysis.** As shown in Table V, the average per-round runtime is about 18.97 seconds. MIST-GNN training takes 17.93 seconds, accounting for about 94.5% of the total time, while SACC and EDGA together take less than 1 second. This shows that the cost of each interaction round is almost equivalent to refitting the risk field, and the overhead of context construction and group assembly is negligible. Fig. 8(b) shows that the account-level LIC remains at the scale of only a few hundred nodes, increasing slightly from about 425 to 449. This is several orders of magnitude smaller than the full graph, which contains about approximately 500K accounts. Therefore, the heterogeneous and temporal modeling of MIST-GNN is applied only to a very small local region of the full network. By constraining the context to a local LIC, SACC reduces the otherwise prohibitive cost of global training to a controlled range.

TABLE V
AVERAGE PER-ROUND RUNTIME OF CLUE2GROUP

| Component | Time(sec) |
|---|---|
| SACC | 0.72 |
| MIST-GNN Training | 17.93 |
| EDGA | 0.27 |
| Clue2Group(Total) | 18.97 |

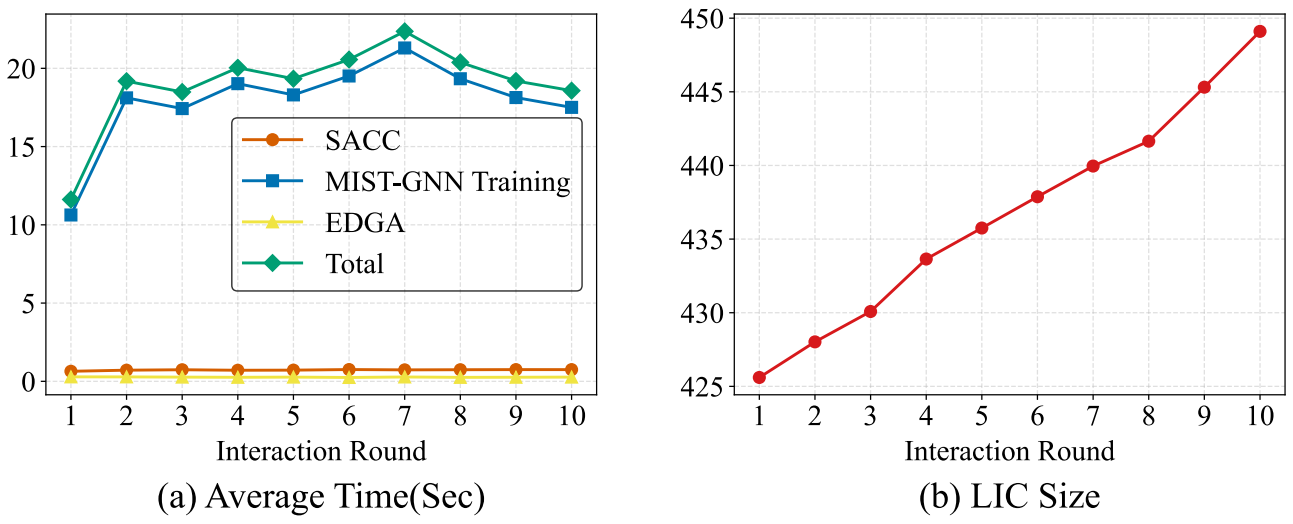

**Fig.8.** Runtime and LIC Size across Interaction Rounds

As shown in Fig. 8(a), the total runtime is dominated by MIST-GNN training. Benefiting from the local context constructed by SACC, the per-round runtime remains stable at around 12–22 seconds. Since each analyst-in-the-loop round also involves analyst inspection and judgment, this latency is acceptable in practice. These results show that Clue2Group can exploit heterogeneous network modeling while still meeting the efficiency requirements of interactive AML investigations.

*3) Case Study*

**Implementation Details**. To illustrate the recovery process of Clue2Group in an interactive investigation, we visualize a case with group ID 333 on HI-Small dataset. Fig. 9 shows the results over 10 interaction rounds. Red nodes denote true group members within the LIC, orange circles denote the group recovered by the model, blue filled nodes denote clue nodes, green nodes denote benign feedback confirmed by the analyst, and gray nodes denote other background nodes.

**Analysis.** As the interaction proceeds, the LIC expands from 31 nodes in round 1 to 74 nodes in round 10, showing that Clue2Group can dynamically enlarge the investigation scope based on newly acquired feedback. In the early rounds, the model recovers a relatively small group around the initial clues. As positive and benign feedback are gradually incorporated, the predicted boundary is progressively refined and expands toward the true group structure. In the middle and later rounds, the predicted group is distributed more consistently around the true members, suggesting that the model jointly exploits risk scores, structural consistency, and prior pattern evidence, rather than relying on a single risk signal.

This case further supports the observation in Fig. 4 that the performance of interactive Clue2Group does not improve monotonically with more rounds, but instead exhibits an optimal interaction window. This suggests that in real scenarios, where obtaining labels is costly, analysts do not need to provide feedback indefinitely. Instead, interaction can stop once recovery performance becomes stable or false positives begin to increase.

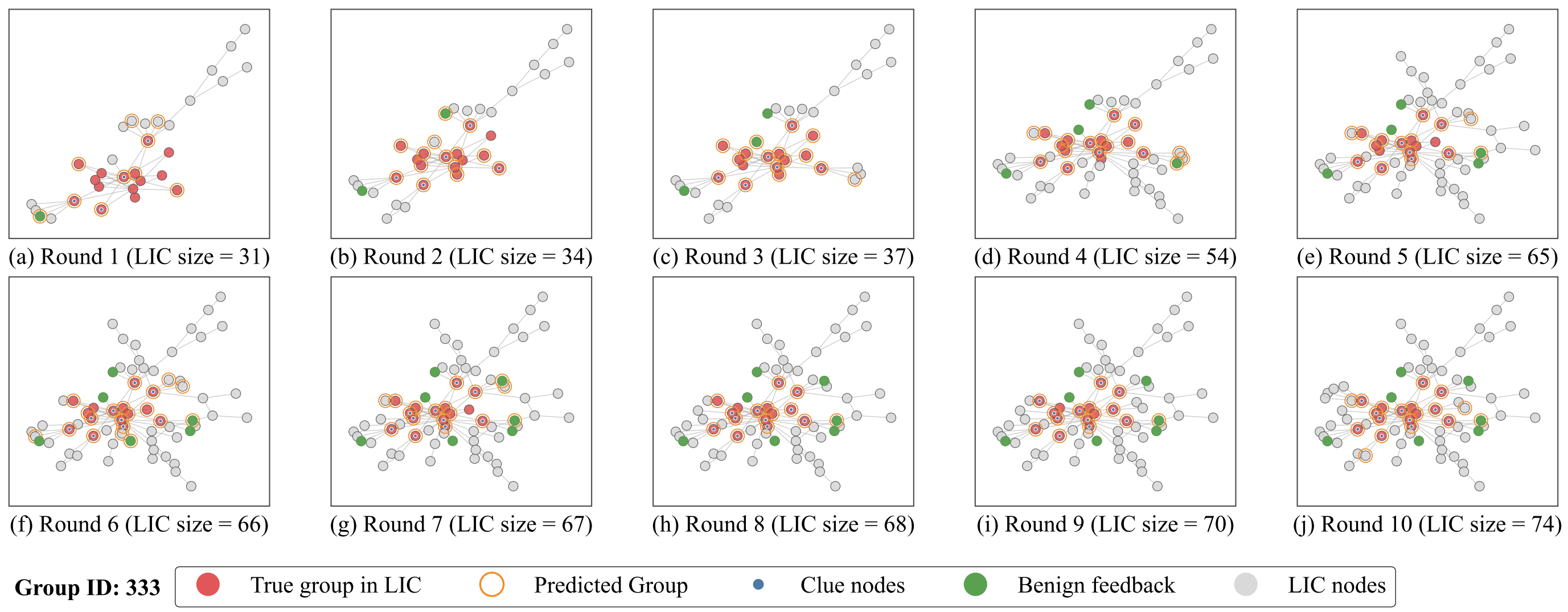

**Fig .9.** Case study of the interactive investigation.

Overall, this case shows that Clue2Group can not only recover the target laundering group, but also provide an interpretable investigation trajectory, supporting clue-guided AML group discovery with limited rounds and high-value feedback.

## VI. Conclusion

This paper studies the problem of clue-guided money laundering group discovery. To address the limitation that existing methods cannot fully support real AML investigation workflows, we formulate Clue-Guided Group Discovery (CGGD) and propose the Clue2Group framework. Experimental results on two large-scale AML benchmarks show that Clue2Group can more accurately recover clue-related laundering groups and effectively support multi-round interactive investigation. Further analysis shows that the framework is robust to changes in clue quantity and clue position, and its computational cost is suitable for practical analyst-in-the-loop scenarios. Overall, Clue2Group can progressively recover target groups from sparse clues. It also provides a more realistic research framework for AML analysis in the investigation stage.

Despite these results, this work has several limitations. First, our experiments are mainly based on public AML benchmarks. Future work should further validate the framework on more real financial networks and more complex laundering scenarios. Second, the current setting assumes that both initial clues and interactive feedback are reliable, without considering noisy clues or mislabeled feedback. Future work can extend Clue2Group to real financial settings with noisy and imperfect interactions.

# Supplementary Material of Clue-Guided Money Laundering Group Discovery

Boyang Wang, Jianing Cao

## I. SACC Time Complexity

**1) Neighborhood expansion.**

For each clue node $c \in C_V$, SACC performs a $k$-hop breadth-first search. Let $d$ be the maximum in-degree/out-degree. The number of nodes visited from one clue node is upper-bounded by $O(d^k)$. Therefore, the complexity for all $|C_V|$ clue nodes is:

$$O\left(|C_V| \cdot d^k\right). \tag{1}$$

**2) Structural enhancement.**

This stage consists of two procedures: path enumeration and fund-flow consistency verification.

**Path enumeration.** For each clue node, SACC enumerates simple paths and cycles within its $k$-hop neighborhood, where the length of each path is bounded by $L_{max}$ and the length of each cycle is bounded by $C_{max}$. Let $|P_c|$ denote the number of paths enumerated from one clue node. The total number of paths over all clue nodes is $|P| = \sum_{c \in C_V} |P_c|$. In the worst case, $|\mathcal{P}_c| = O\left(d^{L^*}\right)$, where $L^* = \max\left(L_{max}, C_{max}\right)$. In practice, most paths are shorter than $L^*$, so the actual enumeration cost is usually much lower than this theoretical upper bound.

**Fund-flow instance construction and consistency verification.** For a path $P_c$ of length $L$, at most $m_{max}$ transaction events are retained for each pair of adjacent accounts. Therefore, the number of fund-flow instances on $P_c$ is upper-bounded by $\left|\mathcal{I}\left(P_c\right)\right| \le \left(m_{\max}\right)^L$. For each fund-flow instance $\mathcal{I}\left(P_c\right)$, SACC computes the coefficient of variation of its amount sequence, with cost $O(L)$. Hence, the verification cost for one path is

$$O\left(\left(m_{max}\right)^L \cdot L\right). \tag{2}$$

In the naive setting, without truncating the number of transactions between account pairs, each path would need to traverse all transaction events. The cost would be $O\left(D^L \cdot L\right)$, where $D$ is the maximum number of transactions between an account pair. In high-frequency transaction networks, $D$ can be as large as hundreds or even thousands. SACC introduces the truncation parameter $m_{max} \ll D$, reducing the exponential base from $D$ to $m_{max}$. This leads to an exponential reduction in the verification cost.

Combining all paths, the total complexity of the structural enhancement stage is

$$O\left(|P| \cdot \left(m_{max}\right)^L \cdot L\right). \tag{3}$$

**3) Subgraph induction.**

SACC merges the candidate node set obtained from neighborhood expansion and structural enhancement, and then induces the corresponding subgraph from the original graph. This step scans the candidate nodes and their associated edges. Its complexity is

$$O\left(|\mathcal{V}_{LIC}| + |\mathcal{E}_{LIC}|\right), \tag{4}$$

which is negligible compared with the previous two stages.

Therefore, the total time complexity of SACC is

$$O\left(|C_V| \cdot d^k + |P| \cdot \left(m_{\max}\right)^{L^*} \cdot L^*\right). \tag{5}$$

## II. Selection of Relevant Historical Transactions for LTCE

Given a transaction event $e = (u, v, t, x)$, we collect candidate neighboring transactions from six relation types:

$$\mathcal{C} = \mathcal{N}^-_{pair}(e) \cup \mathcal{N}^+_{pair}(e) \cup \mathcal{N}^-_{src}(e) \cup \mathcal{N}^+_{src}(e) \cup \mathcal{N}^-_{dst}(e) \cup \mathcal{N}^+_{dst}(e). \tag{6}$$

Here, $\mathcal{N}^-_{pair}(e)$ and $\mathcal{N}^+_{pair}(e)$ denote past and future transactions that involve the same account pair as the current transaction $e$, respectively. The account pair is treated as an unordered pair. That is, two transactions are assigned to the same account-pair group as long as they involve the same two accounts. Similarly, $\mathcal{N}^-_{src}(e)$ and $\mathcal{N}^+_{src}(e)$ denote past and future transactions involving the source account $u$ of the current transaction, while $\mathcal{N}^-_{dst}(e)$ and $\mathcal{N}^+_{dst}(e)$ denote past and future transactions involving the destination account $v$.

Within each relation group, transactions are first sorted by timestamp. If two transactions have the same timestamp, they are further sorted by transaction index. Then, for each transaction, we scan forward and backward within a fixed time window $W$, and retain at most $M$ neighboring transactions in each direction. After collecting candidate neighbors from the six relation types, we merge them into a unified candidate set and remove duplicate transactions. The candidates are then sorted by their absolute temporal distance to the current transaction $e$, $\Delta t_{ee'} = |t_e - t_{e'}|$. Finally, for each transaction $e$, we retain the top-$M$ nearest neighboring transactions:

$$\mathcal{N}(e) = \mathrm{TopM}_{e \in \mathcal{C}}\left(-\Delta t_{ee'}\right). \tag{7}$$

If the number of valid neighboring transactions is smaller than $M$, invalid indices are used for padding. A binary mask $m_{ee'} \in \{0,1\}$ is also constructed for each neighbor position. For each valid neighboring transaction, we further construct a five-dimensional role indicator vector:

$$\mathbf{b}_{ee'} = \left[ b_{ee'}^{src}, b_{ee'}^{dst}, b_{ee'}^{past}, b_{ee'}^{future}, b_{ee'}^{same} \right]. \tag{8}$$

The five dimensions are defined as follows:

$$\begin{aligned} b_{ee'}^{src} &= \mathbf{1}(u_e = u_{e'} \vee u_e = v_{e'}), \\ b_{ee'}^{dst} &= \mathbf{1}(v_e = u_{e'} \vee v_e = v_{e'}), \\ b_{ee'}^{past} &= \mathbf{1}(t_{e'} < t_e \vee (t_{e'} = t_e \wedge e' < e)), \\ b_{ee'}^{future} &= \mathbf{1}(t_{e'} > t_e \vee (t_{e'} = t_e \wedge e' > e)), \\ b_{ee'}^{same} &= \mathbf{1}(u_{e'} = u_e \wedge v_{e'} = v_e). \end{aligned} \tag{9}$$

The first two dimensions indicate whether the neighboring transaction involves the source or destination account of the current transaction. The third and fourth dimensions indicate the temporal direction of the neighboring transaction relative to the current transaction. When two transactions have the same timestamp, their transaction indices are used to determine their temporal order. The last dimension indicates whether the neighboring transaction has exactly the same directed account pair as the current transaction.

In this way, LTCE constructs local temporal context for each transaction from both the account-pair level and the account level. This design captures repeated transfers between the same account pair, as well as related transaction activities around the same account within a short time window. Meanwhile, the role indicator vector preserves the account role and temporal direction of each neighboring transaction, enabling the model to distinguish different types of local transaction patterns.

## III. TRAINING DETAILS

This section provides additional details on the training procedure of MIST-GNN. Since clue-guided money laundering group discovery is conducted at the investigation stage, the model usually has access to only a very small number of confirmed clues. In the initial stage, clue accounts are treated as positive samples, while all other accounts in the LIC are unlabeled. Therefore, the training of MIST-GNN is formulated as a PU learning problem. As the interaction proceeds, analysts may further confirm suspicious accounts or benign accounts. Once confirmed benign accounts are added to the training set, the learning problem becomes a PNU semi-supervised learning setting, where positive, negative, and unlabeled nodes are all available.

### *A. Training Labels and Learning Setting*

For each Local Investigation Context constructed by SACC, let $\mathcal{V}_{LIC}$ denote the set of account nodes, $C_V$ denote the set of clue accounts, and $B_V$ denote the set of confirmed benign accounts. In the initial stage, training labels are provided only by the clue accounts:

$$\mathcal{V}^+ = C_V, \ \ \mathcal{V}^u = V_{LIC} \setminus C_V. \tag{10}$$

Here, $\mathcal{V}^+$ denotes the positive set, and $\mathcal{V}^u$ denotes the unlabeled node set. Since $\mathcal{V}^u$ may contain true laundering group members, we cannot directly treat all unlabeled nodes as negative samples. Instead, we dynamically mine a subset of low-risk nodes from $\mathcal{V}^u$ as pseudo-negative samples to stabilize training.

In the interaction stage, if analysts provide confirmed benign nodes $B_V$, the training sets become:

$$\mathcal{V}^+ = C_V, \ \ \mathcal{V}_c^- = B_V, \ \ \mathcal{V}^u = V_{LIC} \setminus \left( C_V \cup B_V \right), \tag{11}$$

where $\mathcal{V}_c^-$ denotes the set of confirmed negatives. Confirmed negative nodes are not used to construct the pseudo-negative pool. Instead, they serve as explicit supervision signals and are directly included in both the BCE loss and the ranking loss.

### *B. Pseudo-negative Mining*

To reduce label noise in PU learning, we adopt a dynamic pseudo-negative mining strategy. At each epoch, the model produces current risk probability $p_v$.Pseudo-negative samples are selected only from the unlabeled set $\mathcal{V}^u$. By default, we use a stratified pseudo-negative strategy and divide pseudo-negative samples into three groups: **BCE pseudo-negatives, easy negatives, and hard negatives.**

The first group is BCE pseudo-negatives. This set is used for binary cross-entropy training. Specifically, we prioritize unlabeled nodes with low current prediction scores, i.e., nodes below a low-quantile threshold. To avoid incorrectly selecting potential group members close to the clue nodes as negative samples, we can further apply a distance filter. Only nodes whose topological distance to the clue set is at least $d_{min}$ are retained:

$$\mathcal{V}_{\text{bce}}^- = \left\{ v \in \mathcal{V}^u \,\middle|\, p_v \le q_{\text{low}}, \text{dist}\left(v, C_V\right) \ge d_{\text{min}} \right\} \tag{12}$$

The second group is easy negatives. This set is mainly used in the ranking loss. Easy negatives are also selected from nodes with low prediction scores. They represent samples that the model currently considers to be relatively reliable negatives. These nodes provide stable ranking constraints, encouraging positive samples to have higher logits than clearly low-risk nodes.

The third group is hard negatives. Hard negatives are enabled only after the warm-up stage. They are selected from unlabeled nodes whose prediction scores fall within an intermediate range:

$$\mathcal{V}_{\text{hard}}^- = \left\{ v \in \mathcal{V}^u \,\middle|\, q_{\text{hard}}^{\text{low}} \le p_v \le q_{\text{hard}}^{\text{high}} \right\}. \tag{13}$$

These nodes are closer to the current decision boundary of the model. They therefore help improve the model's ability to distinguish boundary cases. To prevent the hard-negative set from becoming too large, we impose a maximum size and randomly subsample the set when it exceeds this limit.

### *C. Weighted Binary Cross-Entropy*

MIST-GNN uses weighted binary cross-entropy as the basic supervised loss. During training, the BCE sample set consists of positive samples, BCE pseudo-negatives, and confirmed negatives:

$$\mathcal{D}_{bce} = \mathcal{V}^+ \cup \mathcal{V}_{bce}^- \cup \mathcal{V}_c^-. \tag{14}$$

The corresponding label for each node is defined as:

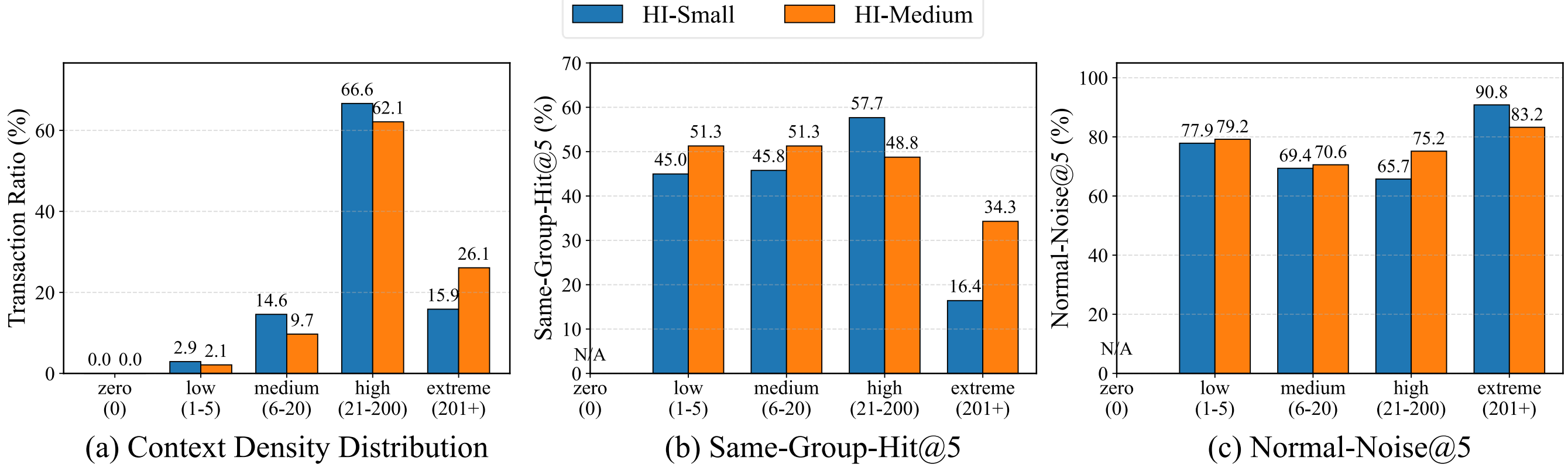

(a) Context Density Distribution (b) Same-Group-Hit@5 (c) Normal-Noise@5

**Fig.1.** This figure compares HI-Small and HI-Medium in terms of local temporal context density and top-5 context quality. The x-axis in all subfigures denotes the number of candidate context transactions within a bidirectional 24-hour window: zero means no candidate transaction, low means 1–5, medium means 6–20, high means 21–200, and extreme means 201 or more. The y-axis in (a) shows the percentage of target laundering transactions in each density range. The y-axis in (b) shows Same-Group-Hit@5, namely whether the top-5 transactions selected by LTCE contain at least one transaction from the same ground-truth laundering group. The y-axis in (c) shows Normal-Noise@5, namely the average proportion of normal transactions among the selected top-5 transactions.

$$y_v = \begin{cases} 1, & v \in \mathcal{V}^+, \\ 0 & v \in \mathcal{V}_{bce}^- \cup \mathcal{V}_c^-. \end{cases} \quad (15)$$

Since different types of samples have different levels of reliability, we assign different weights to the three groups:

$$w_v = \begin{cases} w_+, & v \in \mathcal{V}^+, \\ w_{\text{pn}}, & v \in \mathcal{V}_{bce}^-, \\ w_{\text{cn}}, & v \in \mathcal{V}_c^-. \end{cases} \quad (16)$$

Here, $w_+$, $w_{\text{pn}}$, and $w_{\text{cn}}$ denote the weights for positive samples, pseudo-negative samples, and confirmed negative samples, respectively. Since pseudo-negatives are dynamically mined based on model predictions and distance constraints, they may still contain potential laundering group members.

Therefore, $w_{\text{pn}}$ is usually set lower than the weights of reliable labeled samples to reduce the impact of label noise.

The final weighted BCE loss is defined as:

$$\mathcal{L}_{BCE} = -\frac{1}{|\mathcal{D}_{\text{bce}}|} \sum_{v \in \mathcal{D}_{\text{bce}}} w_v [y_v \log p_v + (1 - y_v) \log(1 - p_v)]. \quad (17)$$

In the initial PU stage, $\mathcal{V}_c^- = \varnothing$. Thus, the BCE loss only uses positive samples and BCE pseudo-negatives. In the interaction stage, analyst-confirmed benign nodes are added to $\mathcal{V}_c^-$ and used as explicit negative supervision during training.

*D. Pairwise Ranking Loss*

Using only the BCE loss may make the model overly dependent on the absolute labels of pseudo-negative samples. To better match the downstream requirement of EDGA, which relies on risk ranking, we further introduce a pairwise ranking loss. This loss encourages clue accounts and confirmed positive accounts to have higher logits than negative or comparison samples.

For a positive sample $i \in \mathcal{V}^+$ and a negative sample $j \in \mathcal{V}^-$, the ranking loss is defined as:

Table I

24-HOUR WINDOW COVERAGE AND TOP-5 CONTEXT QUALITY ON HI-SMALL AND HI-MEDIUM

| Dataset | Max-24h Coverage | Laundering-Hit@5 | Same-Group-Hit@5 | Normal-Noise@5 |
|---|---|---|---|---|
| HI-Small | 100% | **49.02%** | **49.02%** | **70.60%** |
| HI-Medium | 100% | 45.28% | 45.28% | 76.91% |

$$\mathcal{L}_{rank} = \mathbb{E}\left[ m - (\text{logit}_i - \text{logit}_j) \right]_+, \quad (18)$$

where $m$ is the margin. The negative set includes easy negatives, hard negatives, and confirmed negatives in the interaction stage. Specifically, the overall ranking loss is written as:

$$\mathcal{L}_{\text{rank}} = \alpha_e \mathcal{L}_{\text{easy}} + \alpha_h \mathcal{L}_{\text{hard}} + \alpha_c \mathcal{L}_{\text{confirm}}. \quad (19)$$

Here, $\mathcal{L}_{\text{easy}}$ is computed using easy negatives, $\mathcal{L}_{\text{hard}}$ is computed using hard negatives, and $\mathcal{L}_{\text{confirm}}$ is computed using analyst-confirmed negative samples. In the PU stage, no confirmed negatives are available. Therefore, $\mathcal{L}_{\text{confirm}}$ is set to zero, and only easy and hard pseudo-negatives are used to impose ranking constraints. In the PNU stage, confirmed negatives provide more reliable supervision and are explicitly included in the ranking loss.

To reduce the variance caused by random negative sampling, we perform multiple Monte Carlo samples over pseudo-negatives and average the resulting ranking losses. For confirmed negatives, since their number is usually small and their labels are reliable, we directly compute all positive-confirmed negative pairs without negative sampling.

*E. Warm-up Training*

The training process consists of a warm-up stage and a main training stage. The purpose of warm-up is to avoid introducing hard negatives too early, when the model predictions are still unstable. This reduces the impact of incorrectly mined pseudo-negative samples.

During the first $T_w$ epochs, the model is trained only with the weighted BCE loss and the semantic consistency loss:

$$\mathcal{L}^{\text{warm}} = \mathcal{L}_{BCE} + \lambda_2 \mathcal{L}_{con}. \tag{20}$$

In this stage, the ranking loss is disabled, and hard negatives are not used for training. After the warm-up stage, the model enters the main training stage, where both the ranking loss and hard-negative mining are enabled.

### *F. Overall Objective*

In the main training stage, the total loss is defined as:

$$\mathcal{L} = \mathcal{L}_{\text{BCE}} + \lambda_1 \mathcal{L}_{rank} + \lambda_2 \mathcal{L}_{con}. \tag{21}$$

To avoid an overly strong ranking constraint in the later training stage, we decay $\lambda_1$ after a specified epoch:

$$\lambda_1(t) = \begin{cases} \lambda_1, & t < T_d, \\ \lambda_1 \cdot \rho, & t \ge T_d. \end{cases} \tag{22}$$

Where $T_d$ denotes the epoch at which the ranking-loss decay starts, and $\rho$ is the decay factor.

## IV. Density-Dependent Effectiveness of LTCE

In this paper, LTCE uses a fixed 24-hour time window and selects the top-5 historical transactions with the smallest temporal distance among transactions sharing accounts as the local temporal context. To explain the different contributions of LTCE across datasets, we first report its overall context quality in Table I. Max-24h Coverage measures the maximum proportion of laundering transactions in a group that can be covered by any 24-hour window. Laundering-Hit@5 indicates whether the selected top-5 historical transactions contain at least one laundering transaction. Same-Group-Hit@5 further requires this transaction to belong to the same true laundering group as the target transaction. Normal-Noise@5 measures the average proportion of normal transactions in the selected top-5 context.

Both datasets achieve 100% Max-24h Coverage, suggesting that laundering transactions do exhibit local temporal concentration. Therefore, the different effects of LTCE across datasets are not caused by a complete failure of the 24-hour window. Instead, they mainly come from the quality of the selected top-5 historical transactions. Specifically, HI-Small has higher Laundering-Hit@5 and Same-Group-Hit@5 than HI-Medium, while its Normal-Noise@5 is lower. This indicates that, in HI-Small, same-group laundering transactions are more likely to appear among the nearest historical transactions of the target transaction, with less interference from normal transactions. In contrast, although HI-Medium also shows local temporal patterns, it contains a higher density of irrelevant transactions within short time periods. As a result, the top-5 transactions selected in HI-Medium are more likely to include normal-transaction noise, which limits the benefit of LTCE.

Fig. 1 further supports this conclusion from the perspective of context density. Fig. 1(a) shows that most target laundering transactions fall into the high-density range within the bidirectional 24-hour window, indicating that LTCE usually has sufficient context. Fig. 1(b) and Fig. 1(c) show that, in the high-density range where most samples are located, Same-Group-Hit@5 on HI-Small is about 18.2% higher than that on HI-Medium, while Normal-Noise@5 is about 12.6% lower. This suggests that the nearest top-5 historical transactions in HI-Small preserve more same-group laundering signals and contain less normal-transaction noise than those in HI-Medium.

These results also reveal the limitation of the current LTCE design. LTCE uses a fixed 24-hour window and a fixed top-5 context size, and mainly selects historical transactions by temporal distance. When local transaction density is high or normal-transaction noise is strong, the nearest transactions are not necessarily the most informative group context. Future work can therefore explore adaptive time windows and more selective context construction mechanisms.

Overall, Table I and Fig. 1 show that local temporal modeling in LTCE is reasonable, but its effectiveness depends on the quality of the selected top-5 context. For more complex or noisier datasets, more flexible temporal context modeling is needed.

TABLE II

PRECISION AND F1 OF LIC CONSTRUCTION METHODS UNDER DIFFERENT CLUE POSITIONS AND K-HOP.

| k-hop | method | HI-Small | | | | HI-Medium | | | |
|---|---|---|---|---|---|---|---|---|---|
| | | **Precision** | | **F1** | | **Precision** | | **F1** | |
| | | Core | Peripheral | Core | Peripheral | Core | Peripheral | Core | Peripheral |
| k=1 | NE | 0.5763 | 0.5215 | 0.3751 | 0.2644 | **0.5477** | **0.4359** | 0.3415 | 0.2209 |
| | NE+ch | 0.5446 | 0.4898 | 0.4509 | 0.3403 | 0.4822 | 0.4082 | **0.4083** | **0.2883** |
| | NE+cy | **0.5877** | **0.5318** | 0.3963 | 0.2838 | **0.5477** | **0.4359** | 0.3415 | 0.2209 |
| | SACC | 0.5479 | 0.4931 | **0.4572** | **0.3466** | 0.4822 | 0.4082 | **0.4083** | **0.2883** |
| k=2 | NE | 0.2920 | 0.3273 | 0.3758 | 0.4034 | 0.2304 | 0.2470 | 0.3050 | 0.3010 |
| | NE+ch | 0.3026 | 0.3344 | 0.4023 | 0.4255 | 0.2407 | 0.2540 | 0.3308 | 0.3203 |
| | NE+cy | 0.2962 | 0.3316 | 0.3827 | 0.4102 | 0.2304 | 0.2470 | 0.3050 | 0.3010 |
| | SACC | **0.3040** | **0.3359** | **0.4048** | **0.4279** | **0.2407** | **0.2540** | **0.3308** | **0.3203** |
| k=3 | NE | 0.0616 | 0.0807 | 0.1031 | 0.1299 | 0.0422 | 0.0554 | 0.0742 | 0.0936 |
| | NE+ch | 0.0650 | 0.0831 | 0.1094 | 0.1344 | 0.0453 | 0.0578 | 0.0800 | 0.0983 |
| | NE+cy | 0.0621 | 0.0813 | 0.1041 | 0.1309 | 0.0422 | 0.0554 | 0.0742 | 0.0936 |
| | SACC | **0.0652** | **0.0833** | **0.1097** | **0.1347** | **0.0453** | **0.0578** | **0.0800** | **0.0983** |

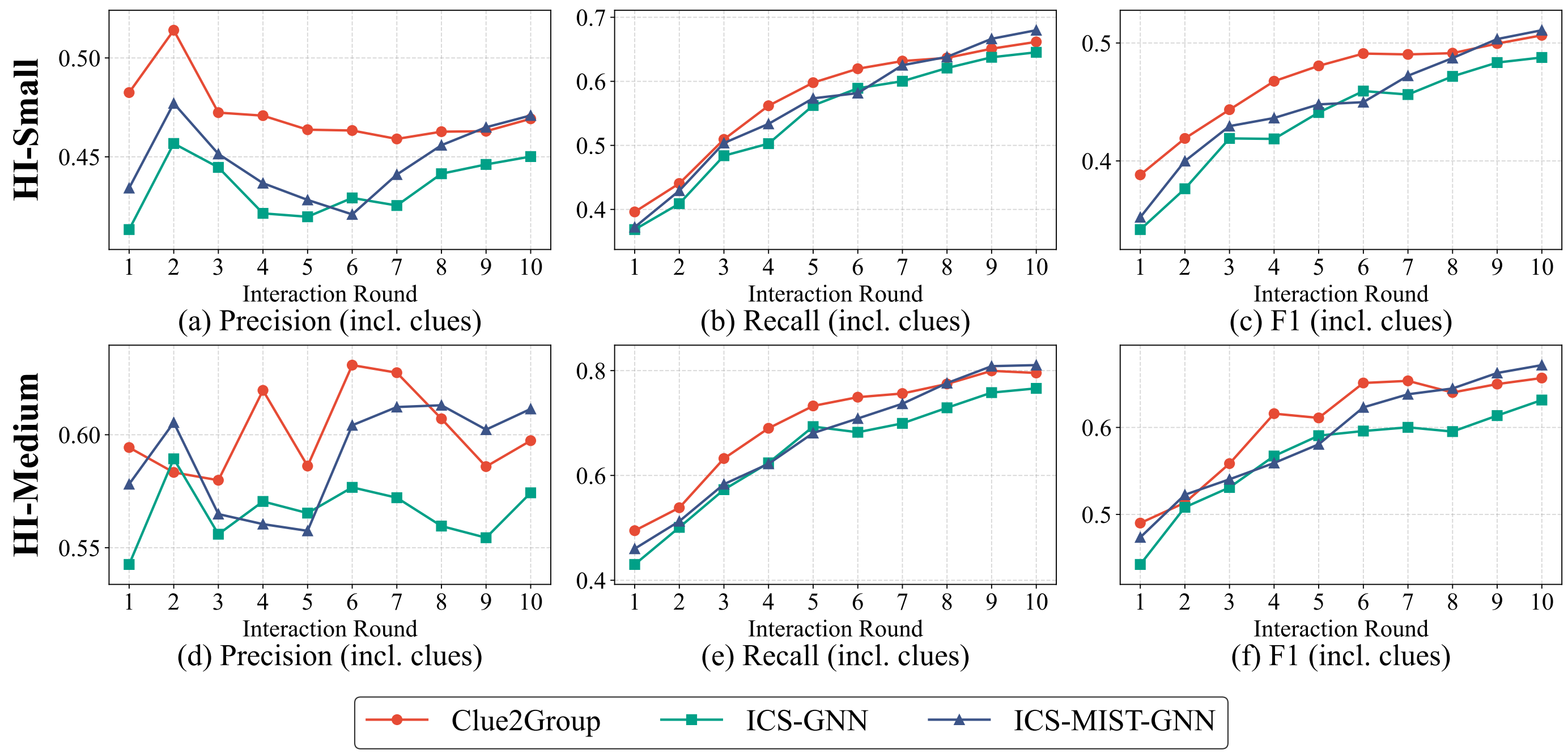

**Fig.2.** Interactive group recovery performance under including-clue evaluation.